\documentclass{article} 
\usepackage[preprint]{colm2026_conference}

\usepackage{microtype}
\usepackage{hyperref}
\usepackage{url}
\usepackage{booktabs}
\usepackage{xspace}
\usepackage{graphicx}
\usepackage{subcaption}
\usepackage{wrapfig}
\usepackage{placeins}
\usepackage[most]{tcolorbox}

\usepackage{lineno}

\definecolor{darkblue}{rgb}{0, 0, 0.5}
\hypersetup{colorlinks=true, citecolor=darkblue, linkcolor=darkblue, urlcolor=darkblue}
\newcommand{\method}{Prefix Sliding\xspace}

\title{\method for efficient test-time scaling}

\newcommand{\aspace}{\hspace{0.8em}}

\definecolor{stanford}{HTML}{8C1515}       
\definecolor{ucsb}{HTML}{C69214}           
\definecolor{uw}{HTML}{4B2E83}             
\definecolor{primeintellect}{HTML}{000000} 

\newcommand{\stanford}{\hspace{.1em}$^{\color{stanford}\boldsymbol{s}}$}
\newcommand{\ucsb}{\hspace{.1em}$^{\color{ucsb}\boldsymbol{c}}$}
\newcommand{\uw}{\hspace{.1em}$^{\color{uw}\boldsymbol{w}}$}
\newcommand{\primeint}{$^{\color{primeintellect}\boldsymbol{p}}$}

\author{%
Niklas Muennighoff\stanford \aspace
Zhengyang Wang\ucsb \aspace
Zeyi Chen\uw \aspace
Weijia Shi\uw \aspace
\textbf{Binyuan Hui} \\
\textbf{John Yang}\stanford \aspace
\textbf{Dapeng Jiang}\uw \aspace
\textbf{Mika Senghaas}\primeint \aspace
\textbf{Fares Obeid}\primeint \aspace
\textbf{Johannes Hagemann}\primeint \\
\textbf{Sami Jaghouar}\primeint \aspace
\textbf{Ludwig Schmidt}\stanford \aspace
\textbf{Percy Liang}\stanford \aspace
\textbf{Jason Wei} \aspace
\textbf{Andrew Y. Ng}\stanford \\
\textbf{Luke Zettlemoyer}\uw \aspace
\textbf{Yejin Choi}\stanford \aspace
\textbf{Mike Lewis}\uw \\
\stanford{}Stanford University \aspace
\ucsb{}University of California at Santa Barbara \aspace
\primeint{}Prime Intellect \\
\uw{}University of Washington \\
\hspace{.1em}
{\tt \href{mailto:n.muennighoff@gmail.com}{n.muennighoff@gmail.com}}
}

\newcommand{\displaycode}[1]{%
\begin{tcolorbox}[
enhanced,
frame hidden,
arc=10pt,
left=0.5cm,
right=0.5cm,
top=0.5cm,
bottom=0.5cm,
grow to left by=1.5pt,
grow to right by=1.5pt,
before skip=\medskipamount,
after skip=\medskipamount,
fontupper=\ttfamily\small,
]
#1
\end{tcolorbox}
}
\definecolor{s1blue}{HTML}{0177B6}
\colorlet{s1bluelight}{s1blue!50}

\begin{document}

\ifcolmsubmission
\linenumbers
\fi

\maketitle

\begin{abstract}
Test-time scaling uses extra test-time compute to improve performance, such as letting language models reason longer when solving a problem. As models keep the entire reasoning trace in memory via full attention, hard tasks that need long thinking can be prohibitively expensive. However, we find most intermediate reasoning tokens lose importance as the model continues reasoning. This calls into question whether retaining them is worth the cost. Based on this insight, we propose \method, which discards tokens during reasoning that are not part of the prefix or the window of the last few thousand tokens. The prefix has key instructions and tools available to the model, while the most recent tokens are the current reasoning the model is working on. This caps the total memory requirement regardless of how long the model reasons, allowing for efficient long-horizon test-time scaling. Without training, \method can make existing models 3x faster while maintaining performance. Training with \method using reinforcement learning can achieve better performance by enabling scaling to reasoning traces beyond a hundred thousand tokens. Ablations show \method outperforms summarizing intermediate tokens or vanilla sliding window. Our code is at 
\href{https://github.com/Muennighoff/prefix-sliding}{https://github.com/Muennighoff/prefix-sliding}
\end{abstract}

\begin{figure}[h]
\centering
\includegraphics[width=\columnwidth]{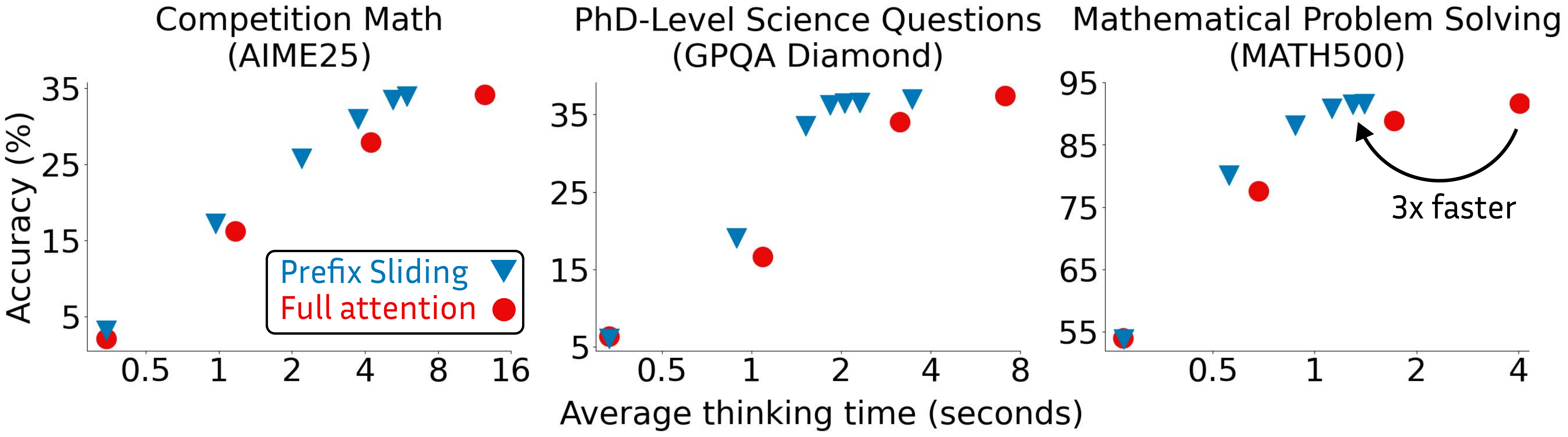}
\caption{\textbf{\method without any training is more efficient than full attention.} \method performs better as it can generate more tokens in the same thinking time than full attention, not because each token it generates is better. Both use the same Qwen3 model (\autoref{sec:setup}). Here, \method uses a window size of 4096; see \autoref{sec:othersw} for other sizes, and \autoref{sec:tab} for the results in tabular form.}
\label{fig:1}
\end{figure}

\newpage

\begin{figure}[t]
\centering
\includegraphics[width=\linewidth]{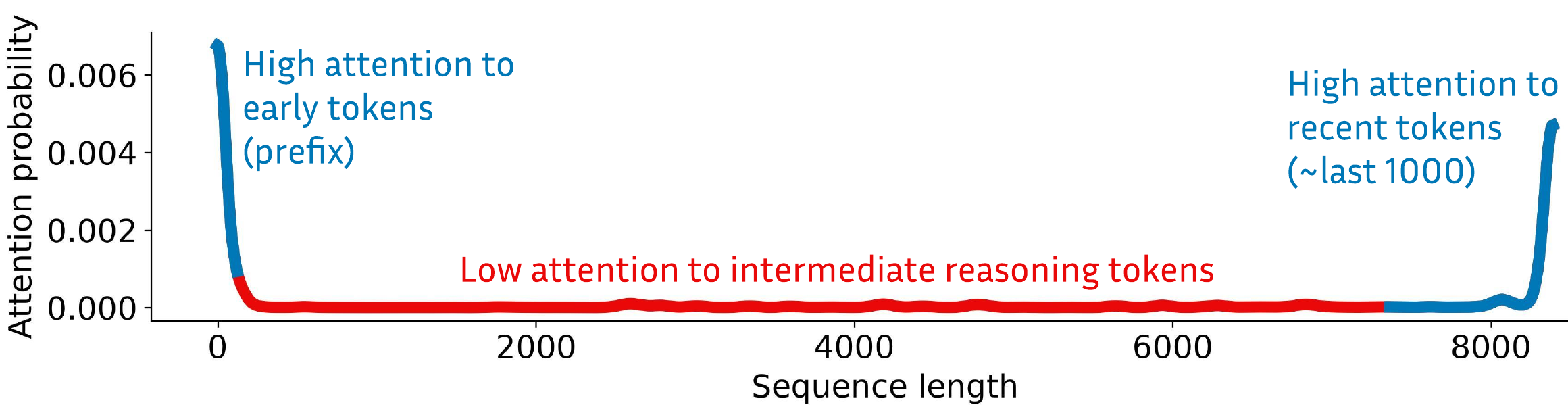}
\caption{\textbf{The first and last few tokens receive most attention during reasoning with full attention.} We plot post-softmax attention probabilities averaged across layers and attention heads in Qwen3-1.7B for an AIME25 reasoning trace. Probabilities are Gaussian smoothed.}
\vspace{-1em}
\label{fig:2}
\end{figure}

\vspace{-.8em}
\section{Introduction}
\label{sec:intro}
\vspace{-.4em}

Test-time scaling improves the performance of language models by using extra compute for hard problems~\citep{o1}. Commonly, this compute is used by letting the model reason longer~\citep{r1,muennighoff2025s1simpletesttimescaling}. However, scaling this approach further is limited by the need to keep the entire reasoning trace in memory via full attention, as used in most language models~\citep{sadhukhan2025kineticsrethinkingtesttimescaling}. With full attention, the cost of each new token grows linearly with the number of already generated tokens, making long context windows prohibitively expensive. Long contexts have more issues, including distraction by old irrelevant tokens~\citep{gema2025inversescalingtesttimecompute}, context poisoning~\citep{gemini25}, repetitive loops~\citep{pipis2025waitwaitwaitreasoning}, and lost knowledge~\citep{liu2023lostmiddlelanguagemodels}.

We explore a simple solution based on two observations. First, intermediate reasoning tokens quickly lose importance. For example, when solving an expression like ``((42 + 84) × 4) - 5'', once the addition 42 + 84 is completed, the reasoning behind that step is no longer needed; only the result matters for the next operation. Second, the prefix and the most recent reasoning tokens, however, are of high importance during generation. The prefix, which includes the system instruction and prompt, contains key information about tools the model can use and the task to complete. It also serves as an ``attention sink'' allowing the model to allocate excess probability weight~\citep{xiao2024efficientstreaminglanguagemodels}. Meanwhile, the most recent reasoning tokens capture what the model is currently working on. This has motivated prior work on letting models generate using a sliding window~\citep{ainslie2020etcencodinglongstructured,gupta2020gmatglobalmemoryaugmentation,beltagy2020longformer,zaheer2021bigbirdtransformerslonger,zhang2025lighttransferlongcontextllmsecretly,fu2025slidingwindowattentiontraining}.

We combine these two observations to propose \method. During reasoning, \method keeps only the prefix and a sliding window in memory. The prefix contains the model instructions. As the model generates, the sliding window advances, removing older intermediate tokens. For example, a 40-token system instruction and a 60-token task prompt could constitute a 100-token prefix. With a 4096-token sliding window, at most 4196 tokens are then kept in memory. The cost of generating an additional token is the same regardless of whether the model has already generated millions or billions of tokens. Such constant cost is necessary to enable very long-horizon test-time scaling.

Empirically, \method can match full-attention performance while running 3$\times$ faster without training, and enables reinforcement learning rollouts beyond 100,000 tokens.

\begin{figure}[t]
\centering
\includegraphics[width=\columnwidth]{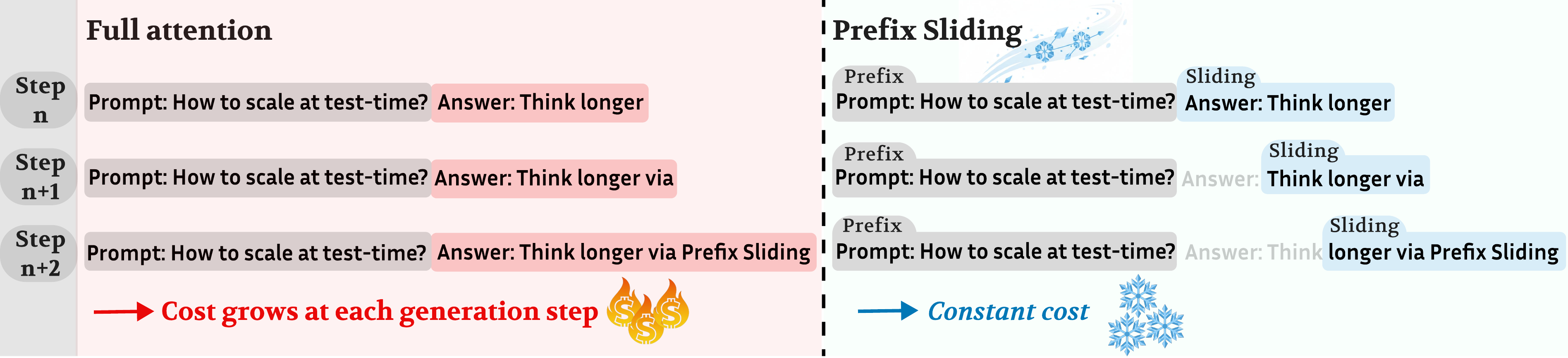}
\caption{\textbf{\method enables efficient long-horizon test-time scaling while full attention inevitably becomes prohibitively expensive.} For \method, the cost at each generation step is constant as attention is only paid to the prefix and sliding window tokens. For full attention, the cost grows at each step due to more tokens in the attention window.}
\label{fig:3}
\end{figure}

\vspace{-.8em}
\section{\method}
\label{sec:method}
\vspace{-.4em}

\paragraph{Motivation} In \autoref{fig:2}, we depict the two observations from \autoref{sec:intro}: (1) Intermediate tokens lack importance and (2) the prefix and recent tokens are key. For the prefix, much probability mass falls on the first four tokens as they function as attention sinks~\citep{xiao2024efficientstreaminglanguagemodels}. The other tokens in the prompt also receive more attention than later intermediate reasoning tokens. The common \verb|<think>| delimiter~\citep{r1} marking the start of the reasoning trace also receives high attention, likely because of its ongoing important role in signaling to the model that it is in thinking mode. The attention probabilities then drop throughout the reasoning trace, but increase sharply toward the end, especially for the token preceding the one being generated.

\begin{wrapfigure}{r}{0.34\textwidth}
\vspace{-1.5em}
\centering
\includegraphics[width=\linewidth]{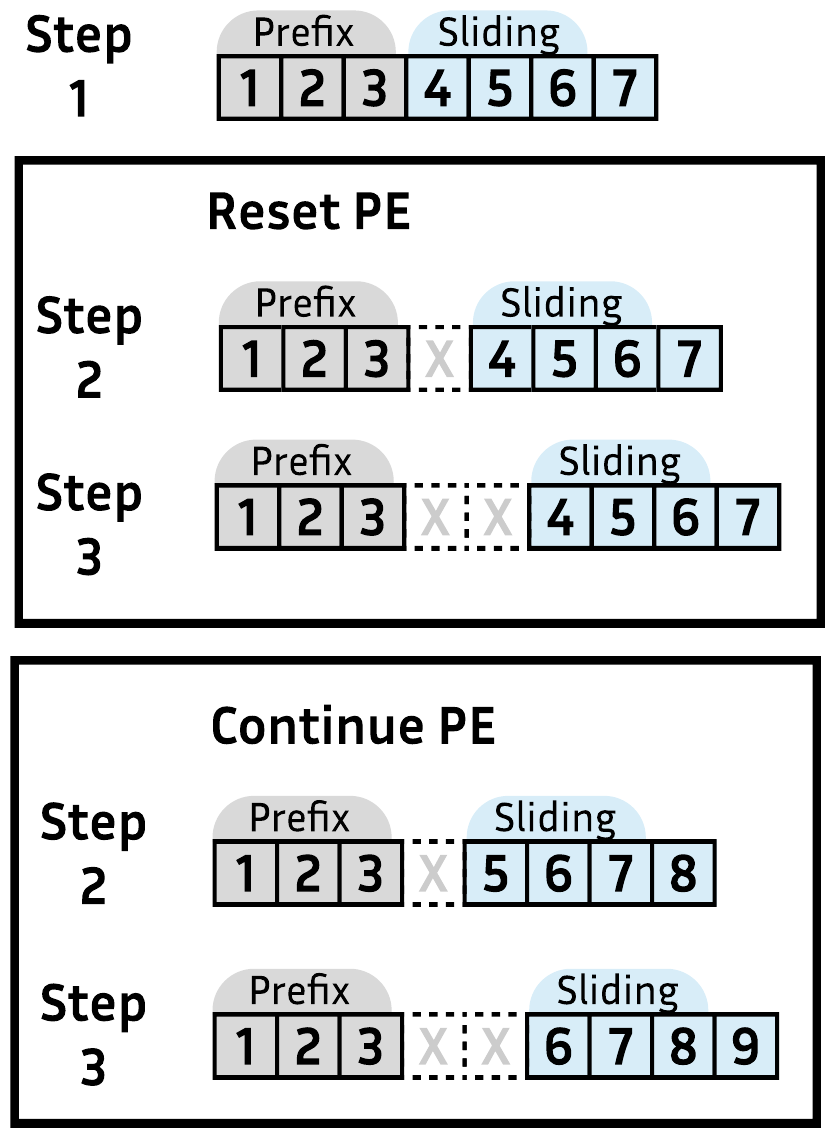}
\caption{\textbf{Position embeddings (PE) with \method.}}
\label{fig:4}
\vspace{-2em}
\end{wrapfigure}

\paragraph{\method without training} \autoref{fig:3} shows how \method works by simply retaining the prefix of tokens and a sliding window. This makes it applicable to generative language models out of the box without any further training. If the model has been trained with position embeddings (PE), such as RoPE~\citep{su2023roformerenhancedtransformerrotary}, then \autoref{fig:4} shows two options for handling them. Compared to Reset PE, Continue PE is more efficient as it does not require reapplying new position embeddings to the same token, but allows reusing cached representations with position embeddings already applied to them. While Continue PE may perform worse~\citep{xiao2024efficientstreaminglanguagemodels}, we have found performance differences insignificant in \autoref{sec:pe}; thus, we use Continue PE. For an even simpler alternative, future work may combine \method with DroPE to simply remove the positional embeddings of pretrained models~\citep{gelberg2025extendingcontextpretrainedllms} or train models without position embeddings from scratch~\citep{kazemnejad2023impactpositionalencodinglength}.

\vspace{0.5em}
\begin{wrapfigure}{l}{0.34\textwidth}
\vspace{-1em}
\centering
\includegraphics[width=\linewidth]{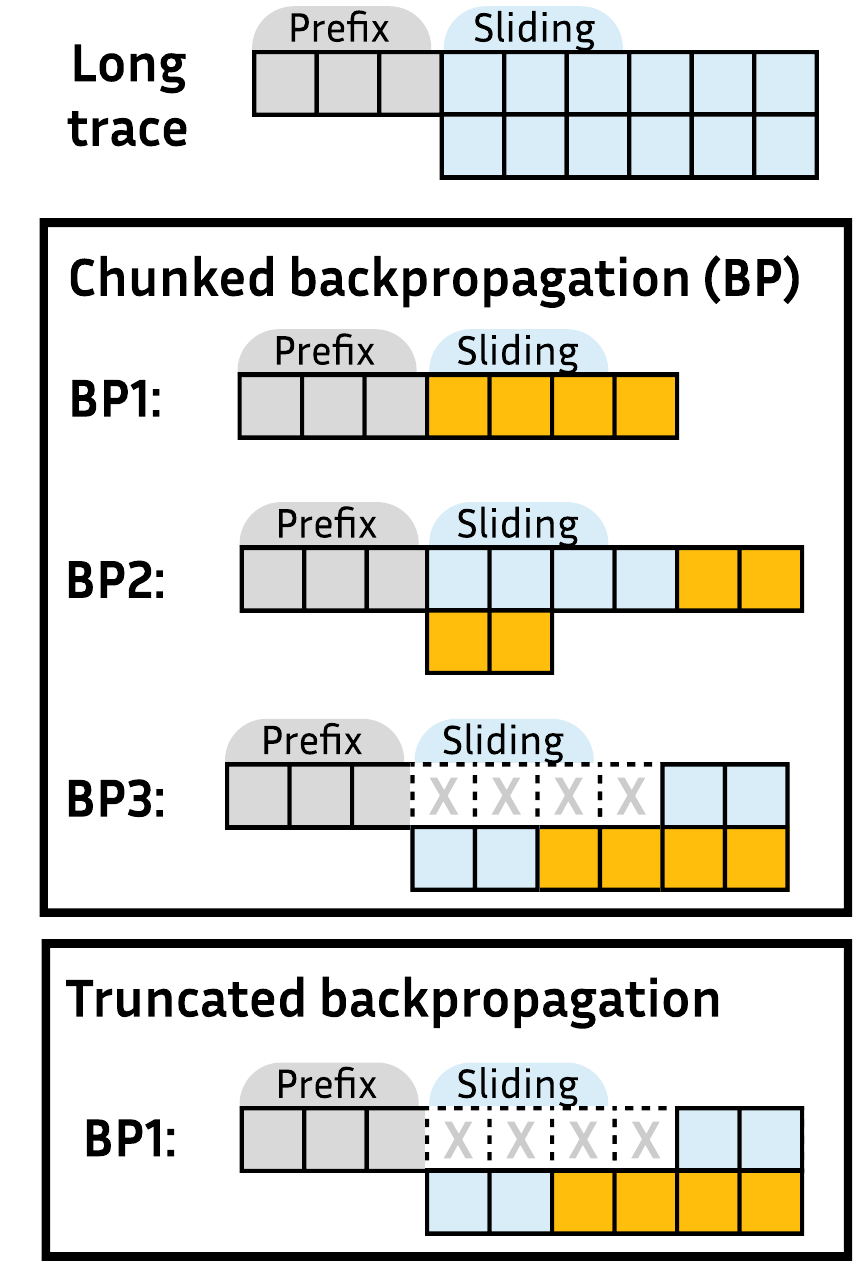}
\caption{\textbf{Backpropagating long reasoning traces with \method.} Yellow marks backpropagated tokens.}
\label{fig:5}
\vspace{-.5em}
\end{wrapfigure}

\paragraph{\method with training} Training with \method enables very long RL rollouts, avoiding the common practice of truncating and discarding overlong generations~\citep{yu2025dapoopensourcellmreinforcement}. Training on completed generations can substantially improve the model. This is best accomplished in an asynchronous RL setup to avoid idle GPUs when other short generations in the same batch are already done~\citep{fu2025areallargescaleasynchronousreinforcement,noukhovitch2025asynchronousrlhffasterefficient}, but also works with synchronous RL. Naive backpropagation of generations that span hundreds of thousands of tokens can lead to out-of-memory errors in the trainer. \autoref{fig:5} depicts two solutions for this issue. Both rely on the limited receptive field of sliding windows. Sliding windows across multiple layers have a theoretical receptive field of $W\times L$, where $W$ is the window size and $L$ the number of layers. However, due to information bottlenecks, it is closer to 1.5$\times W$ in practice~\citep{xiao2025sliding}. Thus, if we want to backpropagate a set of $W$ tokens, we may only need to pass around 1.5$\times W$ of preceding tokens and the prefix to the trainer. Chunked backpropagation backpropagates on a reasoning chain in chunks and accumulates the gradients to ensure near-equivalence with standard full backpropagation. Truncated backpropagation involves only backpropagating on the last chunk. We use \method with truncated backpropagation for our training experiments, as we found its performance can match full attention with full backpropagation in \autoref{sec:trunc}. For example, a model generated a reasoning trace of 100,000 tokens with a sliding window size of 2048. Under truncated backpropagation, we may only send the last 8192 tokens from the sampler to the trainer for gradient computation. We then use the first 6144 tokens only as context and compute the token-level RL loss only on the final 2048 tokens. In our implementation, this is simply a loss mask: the loss for the preceding 6144 tokens is set to zero, and autograd backpropagates normally from the masked loss and only updates with respect to the last 2048 tokens. As the gradients of those 2048 tokens were computed using 4$\times$ the sliding window size, they are very accurate relative to the full 100,000-token generation. Like for \method without training, we also use Continue PE for \method with training. Resetting PE in the trainer is very complex due to teacher-forcing, which is key for training efficiency~\citep{lamb2016professorforcingnewalgorithm,brown2020languagemodelsfewshotlearners}. This is because when resetting PE, each token has seen a different combination of positions before it.

\paragraph{\method kernel implementation}
We implement the \method attention kernel with two-level filtering:

\begin{itemize}
    \item \textbf{Intra-tile masking:} For tiles that partially overlap the allowed attention region (prefix $\cup$ sliding window), we apply an elementwise mask so that only valid (q,k) pairs contribute to the softmax and output. This ensures mathematical correctness without changing the FlashAttention tiling strategy.
    \item \textbf{Inter-tile skipping:} We skip tiles that fall entirely outside the allowed region. Concretely, we restructure the producer--consumer pipeline to iterate over two disjoint block ranges (prefix blocks and window blocks). This avoids redundant loads and computations, substantially matching the efficiency of standard sliding window attention.
\end{itemize}

\section{Setup}
\label{sec:setup}

\paragraph{Modeling} We use the Qwen3-1.7B model unless otherwise specified~\citep{yang2025qwen3technicalreport}. We use vLLM~\citep{kwon2023efficient} with FlashAttention~\citep{dao2022flashattention,dao2023flashattention} for all generations. We write custom kernels for the Nvidia Hopper architecture to enable running \method with FlashAttention. We experiment with sliding window sizes 512, 1024, 2048, 4096, 8192, and 16384. For summary ablations, we treat the summary as a tool call and let the model itself write the summary instead of an external model.

\paragraph{Training} For reinforcement learning experiments, we use GRPO~\citep{shao2024deepseekmathpushinglimitsmathematical} via its synchronous implementation in trl~\citep{vonwerra2022trl}, as well as its asynchronous implementation in prime-rl~\citep{primeintellect2025prime-rl}. We either backpropagate the entire generation or only the last sliding window of tokens, as explained in \autoref{sec:method}. In the latter case, we always pass four times as many of the last tokens to the trainer to ensure we compute accurate gradients for the sliding window. We do not tune other hyperparameters not directly related to \method (e.g., learning rate) and fix them across comparisons. We create our own dataset of math problems and filter it using guessability, verifiability, and difficulty as our three criteria. Details are in \autoref{sec:data}.

\paragraph{Evaluation} We evaluate on standard reasoning benchmarks:  GPQA~\citep{rein2023gpqagraduatelevelgoogleproofqa}, MATH500~\citep{hendrycks2021measuringmathematicalproblemsolving,lightman2023letsverifystepstep}, AIME25~\citep{aime}. We average results across 64 runs to increase confidence in our results, which we refer to as either accuracy or avg@64. We use a temperature of 0.6 and top p of 0.95~\citep{r1}. We verify answers using a small verification library called simpleverify. We use budget forcing to keep generations to specific thinking budgets (without the use of ``Wait'' tokens)~\citep{muennighoff2025s1simpletesttimescaling}. We benchmark models by their average thinking time per sample, measured in seconds, as speed is what users ultimately experience, making it the most important efficiency metric. FLOPs or total generated tokens can be a good proxy, but they miss memory differences among methods.

\newpage

\section{Results}
\label{sec:res}

\begin{wrapfigure}{r}{0.63\textwidth}
\vspace{-4em}
\centering
\includegraphics[width=\linewidth]{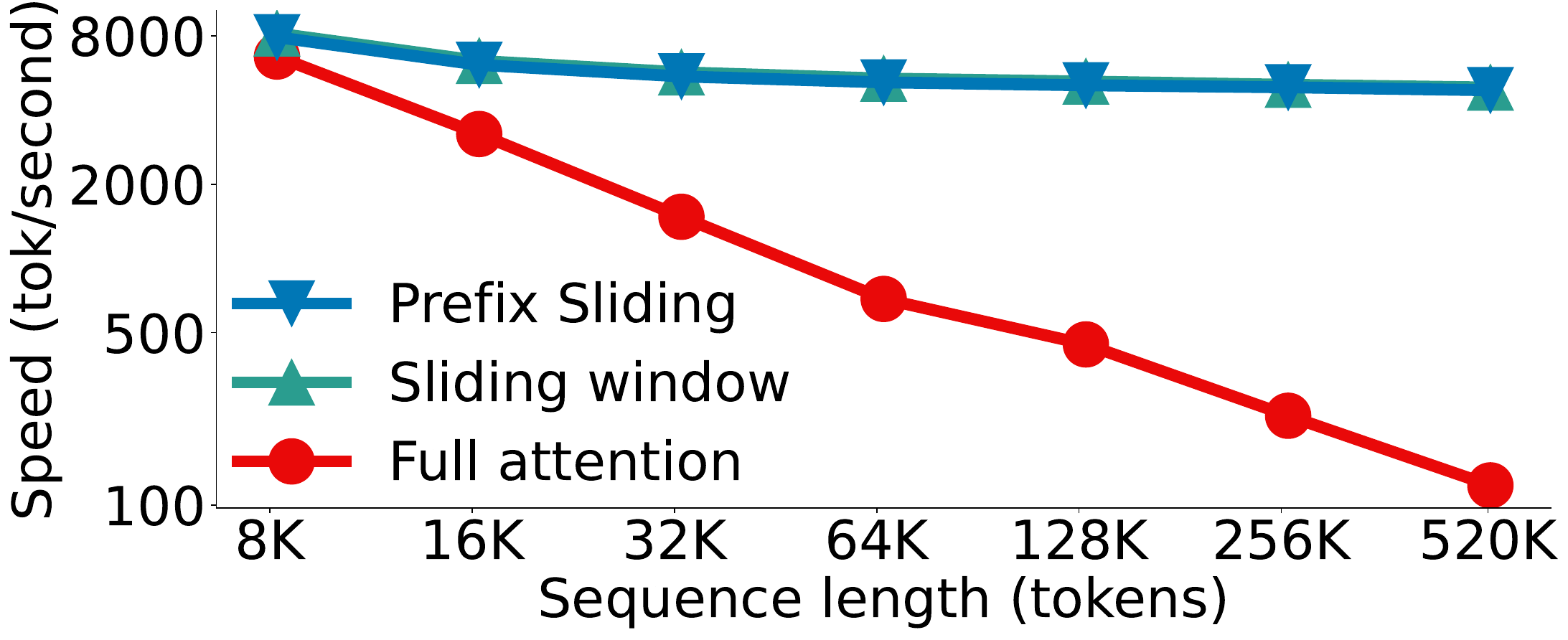}
\vspace{-1.5em}
\caption{\textbf{\method faster than full attention.} We generate 1024 sequences of the respective sequence length with vLLM using its auto batch size, FlashAttention, one 80GB Nvidia H100 GPU, and a window size of 4096 tokens for sliding methods.}
\vspace{-1em}
\label{fig:6}
\end{wrapfigure}

\paragraph{\method without training} \autoref{fig:1} shows \method is more efficient even when applied to an existing model that has been trained with full attention. \autoref{fig:6} shows our FlashAttention kernel for \method reaches about the same speeds as a regular sliding window kernel. A slightly slower speed is expected due to the additional memory requirements of the prefix. The tokens per second for \method and regular sliding window drop initially before stabilizing around 5,000. The initial drop is due to the generation spending less time in the cheap warm-up phase, where the generated tokens are still less than the sliding window. Once the sliding window size is reached, generating each new token costs the same amount. Meanwhile, full attention keeps getting slower indefinitely as the cost per token progressively increases because all prior tokens need to be in memory.

\begin{wrapfigure}{l}{0.52\textwidth}
\vspace{-1.9em}
\centering
\includegraphics[width=\linewidth]{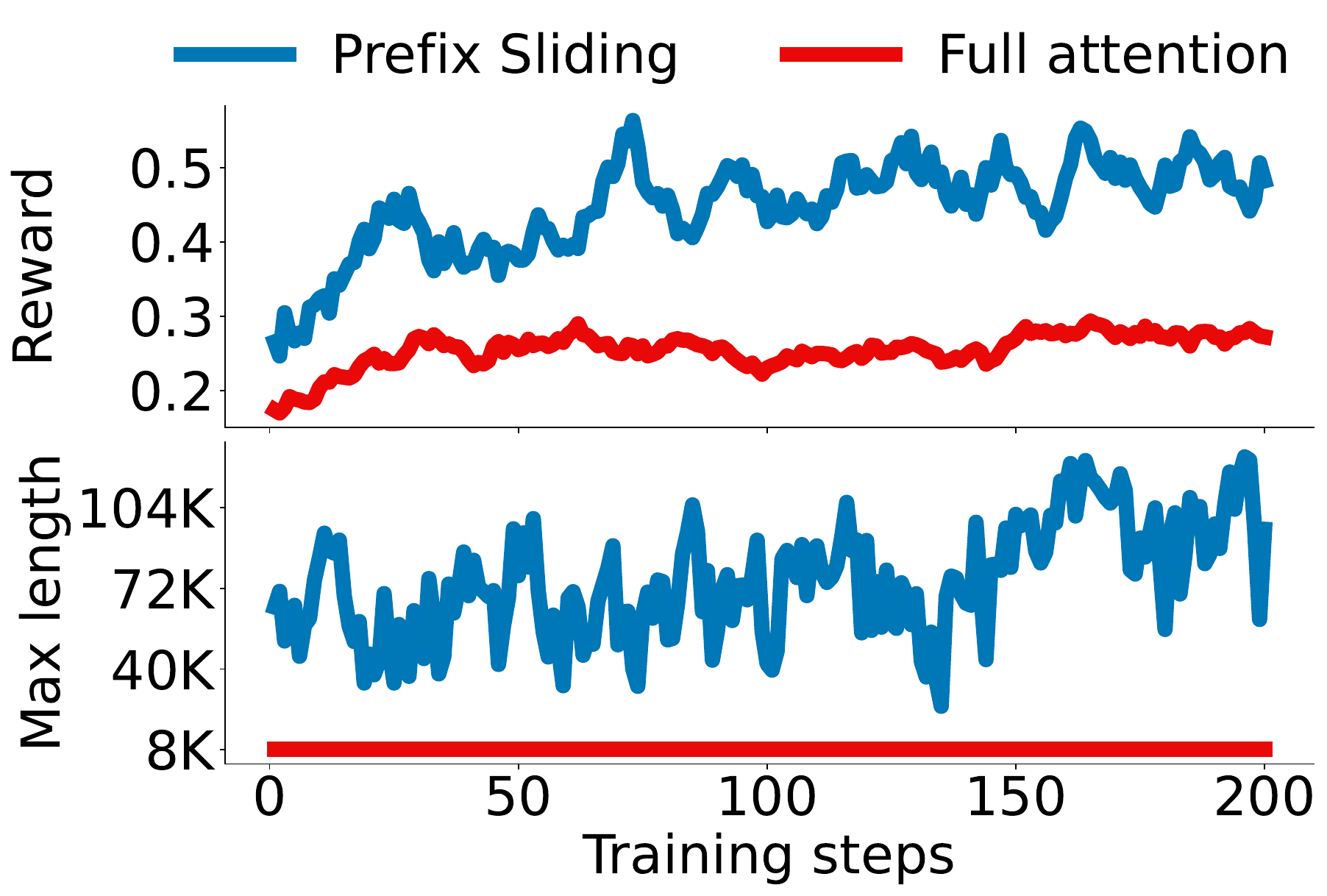}
\vspace{-2em}
\caption{\textbf{\method can improve performance.} We restrict both to near-equal memory budgets: 8,192 max tokens for full attention and an 8,192 sliding window size for \method.}
\label{fig:7}
\vspace{.5em}
\includegraphics[width=\linewidth]{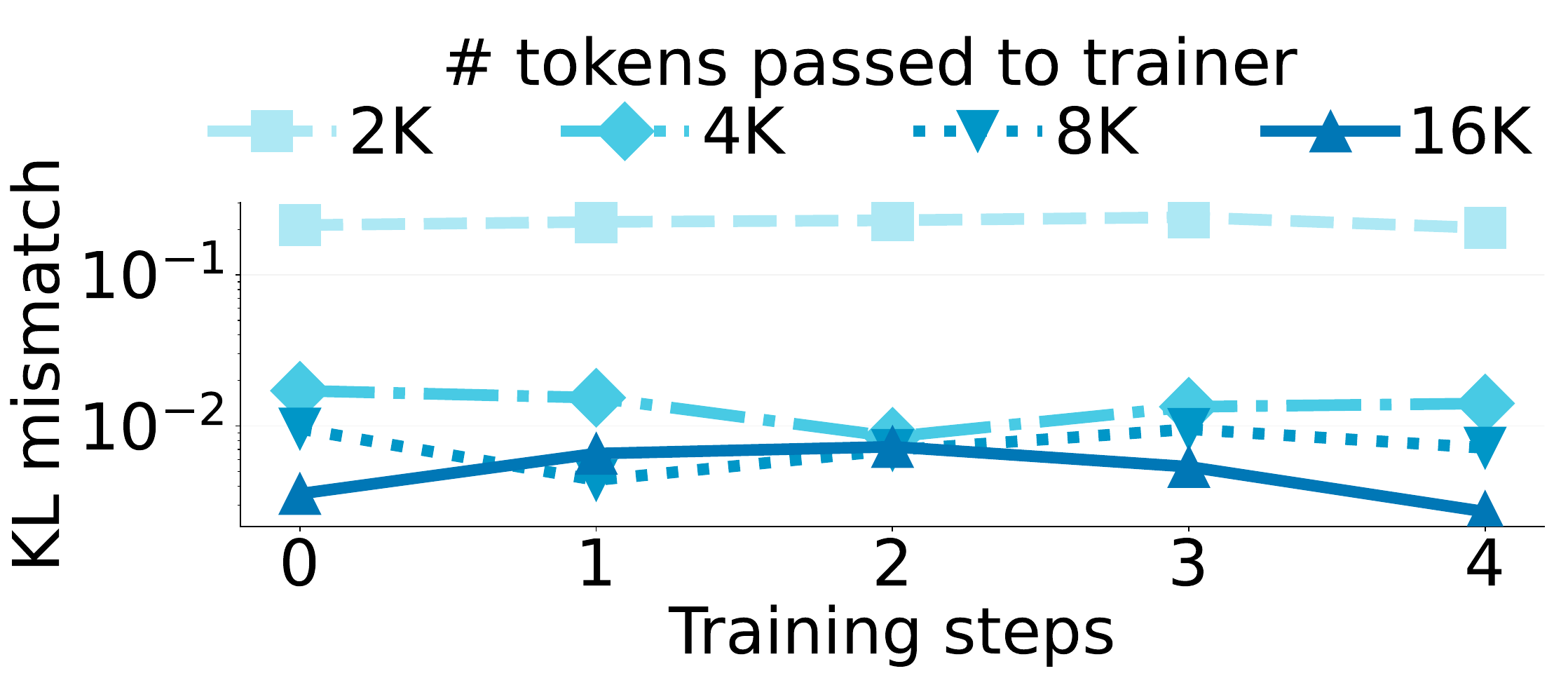}
\vspace{-2em}
\caption{\textbf{Truncated backpropagation numerics.} We compute the Kullback–Leibler (KL) divergence between generator and trainer per-token log probabilities of the last 2048 generated tokens; lower is better. All generate with a max sequence length of 16384 tokens, with a subset passed to the trainer (2K, 4K, 8K, or all 16K). All use \method with a window size of 2048.}
\label{fig:8}
\vspace{-3em}
\end{wrapfigure}

\paragraph{\method with training} \autoref{fig:7} shows that reinforcement learning with \method allows for much longer reasoning traces at near-equal memory budgets, thereby leading to higher rewards. \autoref{fig:8} ablates the choice of tokens passed to the trainer as explained in \autoref{sec:method}. The runs with \method backpropagate one final sliding window (2K tokens). As sliding windows have a limited receptive field (\autoref{sec:method}), only a limited number of tokens prior to this final sliding window are necessary to ensure accurate log probabilities. Only passing the sliding window itself leads to a high KL of above 0.1 as expected. Passing 2$\times$ as much (4K) significantly lowers the mismatch. We go with 4$\times$ (8K) for most runs as it appears to have a slightly lower KL mismatch and is around on par with 8$\times$ (16K). Some remaining KL mismatch is expected due to tiny numerical differences between our custom Flash Attention kernel in the generator and our FlexAttention~\citep{dong2024flexattentionprogrammingmodel} implementation in the trainer. We find the multiplier of 4$\times$ the window size also works well for larger windows. In \autoref{sec:trunc}, we train a 7B model with RL using a window of 8,192 and a multiplier of 4, and find performance comparable to full attention when controlling for sequence length.

\clearpage

\section{Ablations}
\label{sec:abl}

\begin{figure}[t]
\centering
\includegraphics[width=\linewidth]{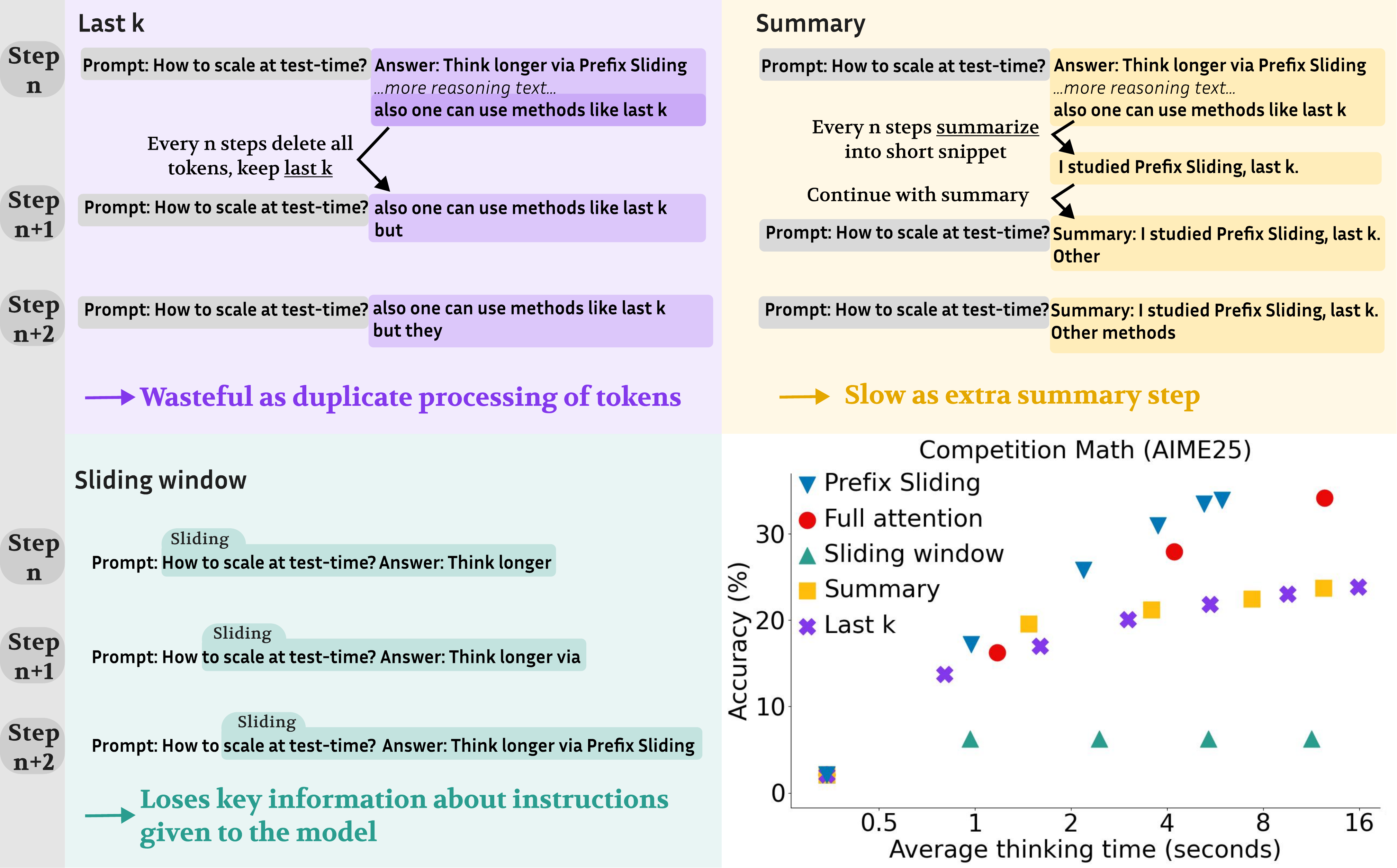}
\caption{\textbf{\method outperforms test-time scaling alternatives.} See \autoref{fig:3} for visual explanations of \method and full attention. For benchmarking on AIME25, we use a maximum generation length of 262144 and a local window of 4096 for all methods except full attention. Thus, after generating 4096 tokens, the window either slides or is restarted with the last k tokens or summary. We set k and the maximum summary length to 256 tokens. See \autoref{sec:setup} and \autoref{sec:othermethods} for additional hyperparameters.}
\label{fig:9}
\end{figure}

We compare \method with three key alternatives, as shown in \autoref{fig:9} and described below. In \autoref{sec:othermethods}, we provide details on their hyperparameter selection and contrast \method with an additional cache-eviction method.

\paragraph{Last k}
\begin{itemize}
    \item \textbf{Explanation}~~~Text is generated until a threshold $n$ is reached, when all text except for the last k tokens is deleted. This way, the context never exceeds the length of the prompt + $n$. As long as $n$ is reasonably small, last k can use full attention without generation inevitably becoming too expensive to continue.
    \item \textbf{Pros and Cons}~~~Last k can be very fast in terms of tokens per second. However, many of those tokens may be wasted. If k is large, the model has to reprocess a lot of tokens. This is because the last k tokens are processed twice: first upon generation and second when their context window changes due to the removal of prior tokens. If k is small, some of the more recent useful tokens may be dropped, and the model may need to regenerate parts of them (e.g., \autoref{fig:lastkissue}). Last k also incurs volatile memory usage; memory usage drops drastically whenever the context is cleared, making it harder to use compute resources optimally.
    \item \textbf{Examples}~~~Variants of last k are often used in agents: For example, after $n$ turns, earlier turns can be deleted, leaving only the most recent few turns (e.g., \citealp{wang2025openhandsopenplatformai}). This approach has also been proposed as "Markovian Thinking / Delethink" in \citet{aghajohari2025markovianthinker}.
\end{itemize}

\clearpage

\paragraph{Summary}
\begin{itemize}
    \item \textbf{Explanation}~~~Text is generated until a threshold $n$ is reached, when all text is summarized, either by the model itself or an external summarizer. Together with the prompt, this summary is then used to start a new context window to continue reasoning. The procedure repeats when $n$ is reached again.
    \item \textbf{Pros and Cons}~~~A benefit of this approach is the model can draw information from anywhere in the current context window for its summary, which in theory could allow it to retain important insights over many summary steps. In practice, however, models struggle to retain important information over many turns~\citep{wang2026lostcompactionevaluatingsideconstraint}. Summary adds complexity by introducing new hyperparameters, such as $n$, the summary length, the summarizer prompt, the summarizer model, and the placement of the summary in the new context. Further, the extra summary-generation step adds overhead, especially if the summary model is large. Like last k, the summary must be processed twice, first upon generation and second when used in the new context window. Its memory usage is also volatile like last k.
    \item \textbf{Examples}~~~This approach has been explored via prompting~\citep{vajipey2025simple}, supervised finetuning~\citep{yan2025inftythinkbreakinglengthlimits,kontonis2026mementoteachingllmsmanage}, or reinforcement learning~\citep{wu2025resumunlockinglonghorizonsearch,yan2026inftythinkeffectiveefficientinfinitehorizon,li2026selfcompactinglanguagemodelagents}. \citet{wu2021recursivelysummarizingbookshuman} use a hierarchical, rather than sequential, version of this approach to handle long inputs. It is also referred to as compaction and used by models like Opus 4.6~\citep{anthropic2026}, GPT 5.4~\citep{singh2025openaigpt5card}, and Composer~\citep{research2026composer2technicalreport}.
\end{itemize}

\paragraph{Sliding window}
\begin{itemize}
    \item \textbf{Explanation}~~~This is a baseline equivalent to \method without prefix.
    \item \textbf{Pros and Cons}~~~The method is very simple. However, as the prefix contains key information about the task, this method performs poorly on longer reasoning tasks. The model forgets which problem it is solving or which tools it can use.
    \item \textbf{Examples}~~~Sliding window attention~\citep{beltagy2020longformer} is common in many large language models, such as gpt-neo~\citep{gpt-neo} and gpt-oss~\citep{openai2025gptoss120bgptoss20bmodel}. To compensate for the lack of information about the prefix, they interleave it with full attention layers that process the entire context.
\end{itemize}

\paragraph{Results} \autoref{fig:9} shows \method provides the best performance efficiency trade-off. \method also adds only one hyperparameter: the size of the sliding window. Pure sliding window attention quickly flattens out due to a lack of information about the task at long thinking times. As soon as the model reaches the sliding window size, it starts losing tokens at the beginning that contain critical information. Last k and summary approaches can reach good performance but are fundamentally constrained by their required token reprocessing and extra summary step. They also add other complexity overhead by requiring more hyperparameters and generation restarts.

\section{Related Work}
\label{sec:rel}

\paragraph{Test-time scaling} Current methods to scale compute at test-time are either \textbf{sequential} or \textbf{parallel}~\citep{snell2024scalingllmtesttimecompute,muennighoff2025s1simpletesttimescaling}. Parallel methods allow for infinite test-time scaling by design, e.g., majority voting~\citep{wang2023selfconsistencyimproveschainthought} simply requires launching more parallel processes to try to solve the same question. However, they face stark diminishing returns~\citep{brown2024largelanguagemonkeysscaling,ehrlich2025codemonkeysscalingtesttimecompute,schaeffer2025largelanguagemonkeyspower}. Sequential scaling can scale better than parallel~\citep{muennighoff2025s1simpletesttimescaling}. While there has been much work on improving sequential scaling and reasoning models in general~\citep{zhang2025surveytesttimescalinglarge,aggarwal2025l1controllinglongreasoning,yue2025doesreinforcementlearningreally,yong2025crosslingualreasoningtesttimescaling,lu2025retrosearchexploringuntakenpaths,guha2025openthoughtsdatarecipesreasoning}, long-horizon scaling remains a fundamental limitation due to the quadratic complexity of the transformer~\citep{vaswani2017attention}. We build a simple method that significantly improves reasoning efficiency while enabling long-horizon scaling due to its constant cost, as elaborated in the next paragraph.

\clearpage

\begin{wrapfigure}{r}{0.5\textwidth}
\centering
\includegraphics[width=\linewidth]{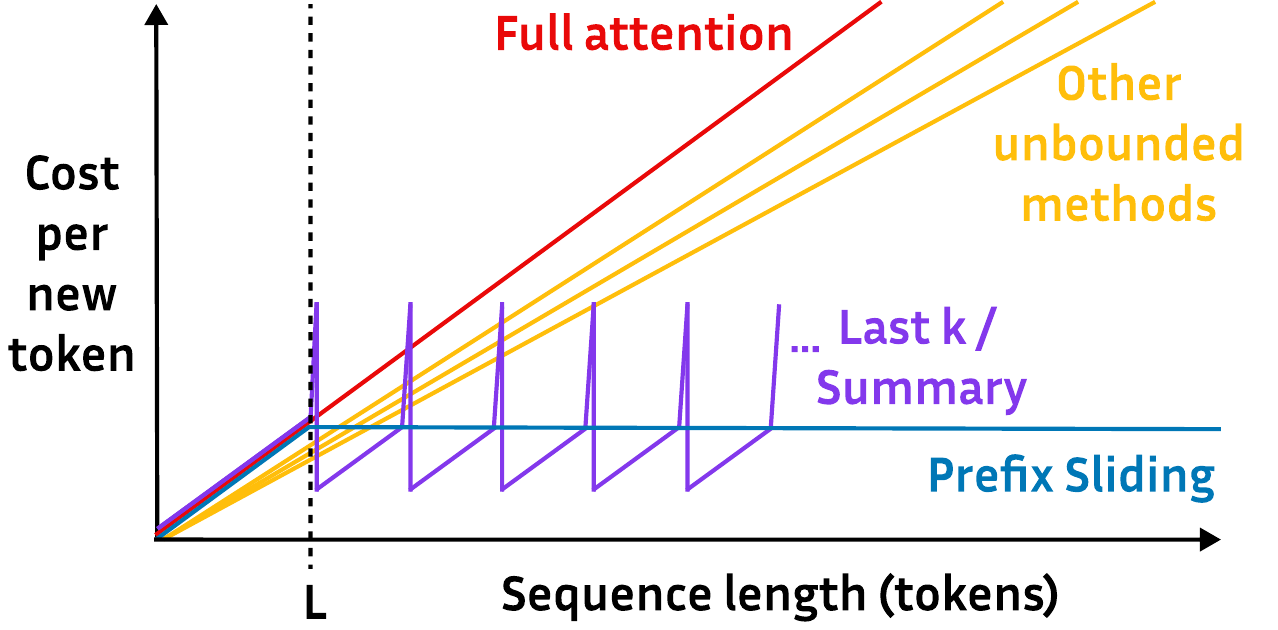}
\caption{\textbf{\method has constant cost per new token in the limit.} ``L'' is where the sequence length reaches the size of the sliding window + the prefix. Last k and summary follow a sawtooth pattern with a cost spike at each chunk end, as the last k tokens need to be passed over or the entire chunk summarized.}
\label{fig:10}
\end{wrapfigure}
\paragraph{Context extension} We distinguish between methods that are \textbf{bounded} and \textbf{unbounded} in their cost per new token. Bounded methods are asymptotically constant; in the limit, they cost at most a certain amount per new token~\citep{peng2022abcattentionboundedmemorycontrol,munkhdalai2024leavecontextbehindefficient,yang2024gatedlinearattentiontransformers}. Unbounded methods cost more for each new token in the limit, even if they may exhibit subquadratic complexity~\citep{child2019generatinglongsequencessparse,kitaev2020reformerefficienttransformer,wang2020linformerselfattentionlinearcomplexity,jaegle2021perceiver,xiong2021nystromformernystrombasedalgorithmapproximating,choromanski2022rethinkingattentionperformers,liu2023ringattentionblockwisetransformers,ho2024blocktransformerglobaltolocallanguage,deepseekai2025deepseekv32pushingfrontieropen,shyam2025treeattentiontopologyawaredecoding}. Full attention is unbounded: Every generated token gets more expensive. One can trade off space and time complexity of full attention to make one bounded, but the other stays unbounded~\citep{rabe2022selfattentiondoesneedon2}. Crucially, to enable infinite test-time scaling, i.e., models that reason for weeks, cost must be bounded per new token. \autoref{fig:10} shows \method is bounded once the sequence length reaches the combined size of the prefix and the sliding window. Other bounded methods include RNNs (e.g. RWKV~\citep{peng2023rwkv,peng2024eagle}), SSMs (e.g. Mamba~\citep{gu2024mambalineartimesequencemodeling,dao2024transformersssmsgeneralizedmodels,gu2022efficientlymodelinglongsequences,wang2024mambabytetokenfreeselectivestate}), and transformer-based approaches~\citep{dai2019transformerxlattentivelanguagemodels,rae2019compressivetransformerslongrangesequence,chevalier2023adaptinglanguagemodelscompress,tandon2025endtoendtesttimetraininglong}. However, they do not work out of the box with existing models but require training models to adapt them. \method works with existing pretrained models without further training and can optionally also be used for training. Last k and summary approaches are also bounded and work out of the box, but exhibit irregular cost as shown in \autoref{fig:10} due to deleting and refilling of the context window. This makes full GPU utilization difficult. They also incur fundamental latency overhead due to duplicate token processing, which \method bypasses, as detailed in \autoref{sec:abl}. One way to view \method is as sliding window attention with global tokens~\citep{beltagy2020longformer}, but with many consecutive global tokens forming a prefix that preserves task instructions and other necessary context for reasoning tasks. Another related approach is StreamingLLM~\citep{xiao2024efficientstreaminglanguagemodels}, which retains only a few fixed initial tokens, e.g., 4.

\section{Conclusion}
\label{sec:con}

We propose \method to enable language models to reason for extremely long horizons. Even at short reasoning horizons of only thousands of tokens, \method is more efficient than the status quo of using full attention. \method is applicable to language models without further training. It can also be used during training with reinforcement learning. It outperforms alternatives that could also support infinite test-time scaling. We hope that enabling language models to think longer via \method inspires future work on solving ever harder problems with language models.

\section*{Limitations}

\paragraph{Limited comparisons} We restrict ourselves to empirical comparisons with alternatives that fulfill two properties: (1) they work on existing pretrained transformers out of the box and (2) they lead to a bounded cost per new token (see \autoref{sec:rel}). This excludes many approaches, such as alternative architectures, subquadratic methods, or mixed sliding window models~\citep{tay2020long,bulatov2022recurrentmemorytransformer,hwang2024transformerfamfeedbackattentionworking,he2025hmthierarchicalmemorytransformer,li2025ccfcontextcompressionframework}. Future work may consider relaxing the first criterion by comparing with alternative architectures that still exhibit a bounded cost per new token, such as RNNs. We consider it beyond the scope of this work, as it likely requires pretraining models from scratch to control for computational resources and other hyperparameters.

\begin{wrapfigure}{r}{0.43\textwidth}
\centering
\includegraphics[width=\linewidth]{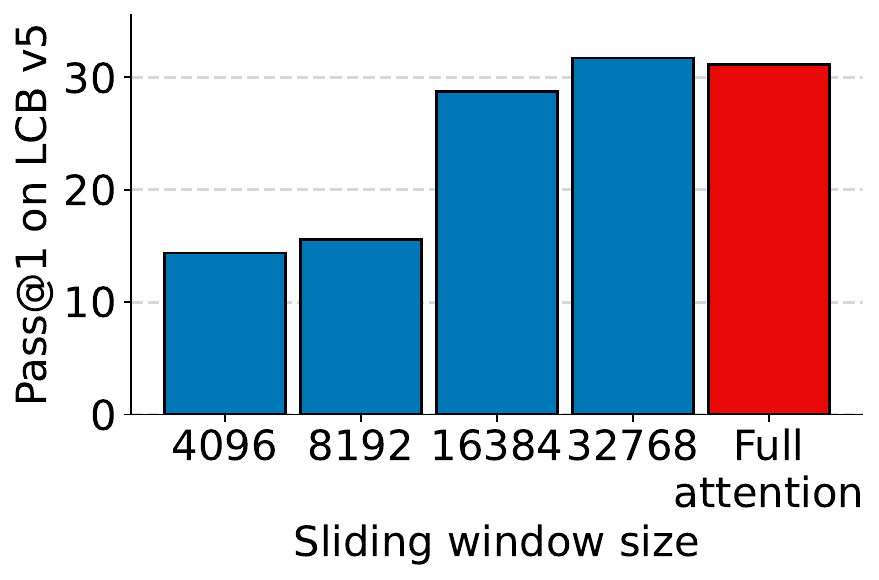}
\caption{\textbf{LiveCodeBench (LCB) requires \method with a window of at least 16384 to match full attention.} Max tokens are 262144.}
\label{fig:11}
\end{wrapfigure}

\paragraph{Information loss} While intermediate tokens can lack importance for later reasoning, as we show in \autoref{sec:method}, sometimes this is not the case. \autoref{fig:11} shows this limitation on the example of LiveCodeBench, where a larger window size is necessary to match full attention. Inspecting samples reveals that the issue is likely that for LiveCodeBench, the model starts a function implementation during reasoning and then thinks using comments for potentially thousands of tokens (see \autoref{fig:lcb} for an example). By the time it continues coding, the beginning of the code may have moved outside its sliding window. This evaluation is without any training. Training with \method during reinforcement learning would likely teach the model to simply adapt its commenting behavior, thus enabling a shorter window size. Alternatively, a mechanism for the model to append sliding tokens to the prefix, or another knowledge store, may avoid the need for larger window sizes.

\begin{wrapfigure}{r}{0.43\textwidth}
\vspace{-1.5em}
\centering
\includegraphics[width=\linewidth]{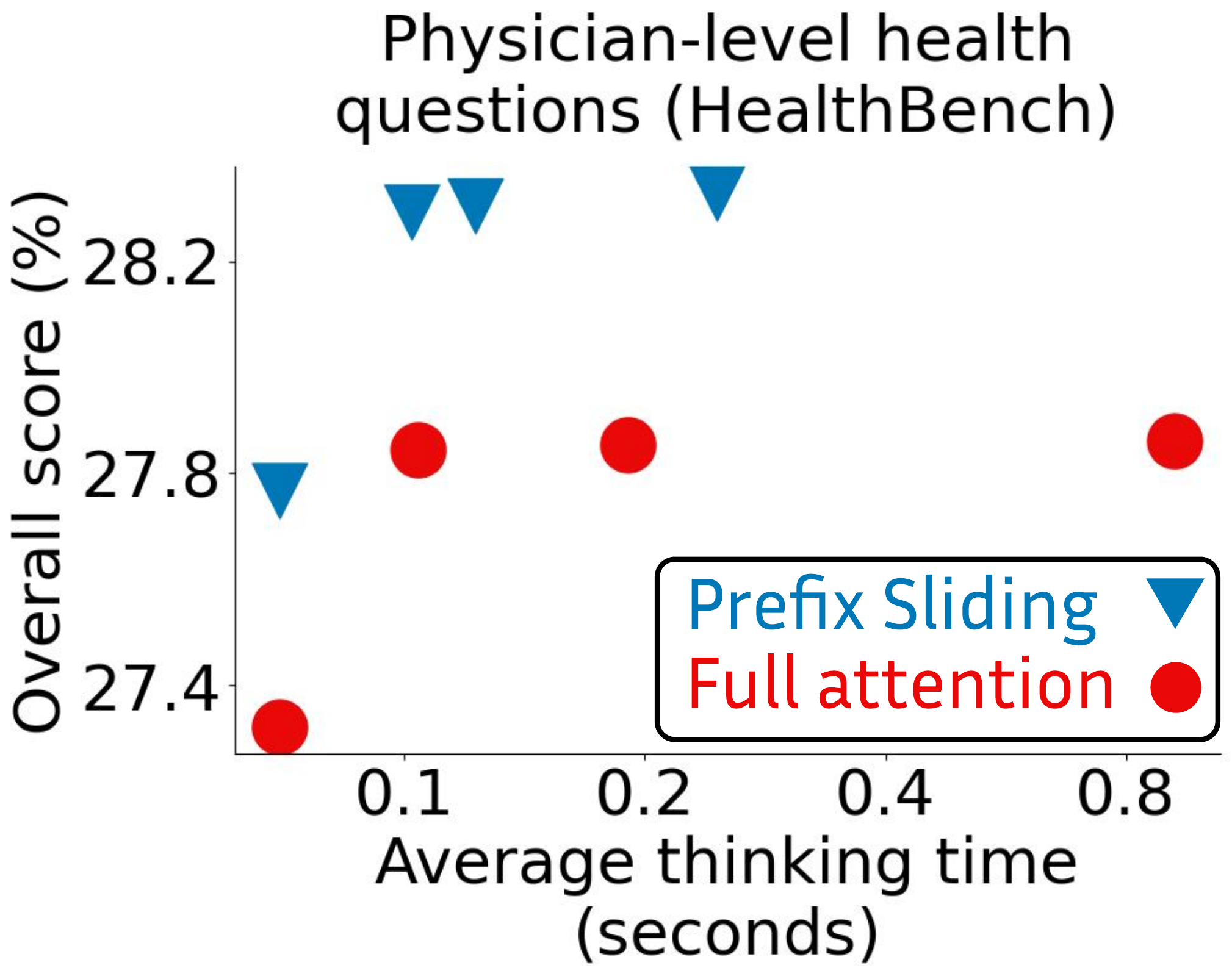}
\vspace{-1.5em}
\caption{\textbf{Fast tasks benefit less from \method.} The sliding window size is 2048 for \method.}
\label{fig:12}
\end{wrapfigure}
\paragraph{Limited benefit for short generations} As is clear from \autoref{fig:6}, the benefits of \method are larger the longer the generation of the model. For short generations, a larger proportion of the generation still uses full attention while the sliding window size has not yet been reached. We call this the sliding window warm-up phase. Only after this phase is complete does the window slide and evict old tokens, thereby offering major speed-ups over full attention. In \autoref{fig:12}, we benchmark \method on a task that requires only 2086 tokens on average: HealthBench~\citep{arora2025healthbenchevaluatinglargelanguage}. As we use a sliding window of 2048 for \method, the model slides very rarely. It is equivalent to full attention for the many samples that require fewer than 2048 tokens. Thus, there is little room for any speed-up.

\paragraph{System outputs and multi-turn} In agentic tasks, a model may read the contents of a website or read files, which could flood the entire context window. This could be problematic because if the sliding window is smaller than the content the model is trying to read, then it strictly cannot read the entire content. Even worse, it may lose important content, as its sliding window is flooded with this new output. A second related issue is what to do with future user instructions in a multi-turn setup. Append them to the prefix? Let the sliding window remove them eventually? These two problems also exist with summarization or last k techniques, assuming the same window size. Extensive reinforcement learning would likely teach the model to be extra careful to avoid this behavior. Another approach could be to let the model learn to read content step by step rather than in one go (e.g., using ``head'' rather than ``cat'' commands in Unix). One can also add automatic guardrails that prevent excessive outputs in the model's context by quickly checking such outputs before and not providing them to the model beyond a prespecified threshold.

\paragraph{Scale} In this work, we scale up \method to hundreds of thousands of thinking tokens and 7 billion parameter models across training-free and reinforcement learning training setups. Future work is necessary to scale up \method further and study its trends.

\section*{Reproducibility Statement}

As \method is very simple and \autoref{sec:method} describes it in detail, it is likely easy to reproduce our key results using only the paper. We also make our code public at \href{https://github.com/Muennighoff/prefix-sliding}{https://github.com/Muennighoff/prefix-sliding}.

\section*{Author Contributions}

Niklas Muennighoff ran training, evaluation, wrote the paper, led the project. Zhengyang Wang, Niklas Muennighoff worked on kernels. Zeyi Chen, Niklas Muennighoff, Dapeng Jiang implemented ablations. Niklas Muennighoff, John Yang, Weijia Shi implemented evaluation. Niklas Muennighoff, Binyuan Hui, John Yang made datasets. Mike Lewis, Yejin Choi, Luke Zettlemoyer, Weijia Shi, Andrew Y. Ng, Jason Wei, Percy Liang, Ludwig Schmidt, Sami Jaghouar, Johannes Hagemann, Fares Obeid, Mika Senghaas advised the project.

\section*{Acknowledgments}

We are extremely thankful to Laude Institute for supporting this work. Research supported by the NVIDIA Academic Grant Program. NM is supported by a graduate fellowship award from Knight-Hennessy Scholars at Stanford University. This work was supported by IITP funded by the Korean Government (MSIT) (No. RS-2024-00457882, National AI Research Lab Project). This research was supported in part by a gift from DSO National Laboratories.

\bibliography{colm2026_conference}

@misc{pipis2025waitwaitwaitreasoning,
      title={Wait, Wait, Wait... Why Do Reasoning Models Loop?}, 
      author={Charilaos Pipis and Shivam Garg and Vasilis Kontonis and Vaishnavi Shrivastava and Akshay Krishnamurthy and Dimitris Papailiopoulos},
      year={2025},
      eprint={2512.12895},
      archivePrefix={arXiv},
      primaryClass={cs.LG},
      url={https://arxiv.org/abs/2512.12895}, 
}

@misc{gelberg2025extendingcontextpretrainedllms,
      title={Extending the Context of Pretrained LLMs by Dropping Their Positional Embeddings}, 
      author={Yoav Gelberg and Koshi Eguchi and Takuya Akiba and Edoardo Cetin},
      year={2025},
      eprint={2512.12167},
      archivePrefix={arXiv},
      primaryClass={cs.CL},
      url={https://arxiv.org/abs/2512.12167}, 
}

@misc{r1,
      title={DeepSeek-R1: Incentivizing Reasoning Capability in LLMs via Reinforcement Learning}, 
      author={DeepSeek-AI and Daya Guo and Dejian Yang and Haowei Zhang and Junxiao Song and Ruoyu Zhang and Runxin Xu and Qihao Zhu and Shirong Ma and Peiyi Wang and Xiao Bi and others},
      year={2025},
      eprint={2501.12948},
      archivePrefix={arXiv},
      primaryClass={cs.CL},
      url={https://arxiv.org/abs/2501.12948}, 
}

@article{k1.5,
  title={Kimi k1. 5: Scaling reinforcement learning with llms},
  author={Kimi, Team and Du, Angang and Gao, Bofei and Xing, Bowei and Jiang, Changjiu and Chen, Cheng and Li, Cheng and Xiao, Chenjun and Du, Chenzhuang and Liao, Chonghua and others},
  journal={arXiv preprint arXiv:2501.12599},
  year={2025}
}

@misc{lightman2023letsverifystepstep,
      title={Let's Verify Step by Step}, 
      author={Hunter Lightman and Vineet Kosaraju and Yura Burda and Harri Edwards and Bowen Baker and Teddy Lee and Jan Leike and John Schulman and Ilya Sutskever and Karl Cobbe},
      year={2023},
      eprint={2305.20050},
      archivePrefix={arXiv},
      primaryClass={cs.LG},
      url={https://arxiv.org/abs/2305.20050}, 
}

@misc{dong2024flexattentionprogrammingmodel,
      title={Flex Attention: A Programming Model for Generating Optimized Attention Kernels}, 
      author={Juechu Dong and Boyuan Feng and Driss Guessous and Yanbo Liang and Horace He},
      year={2024},
      eprint={2412.05496},
      archivePrefix={arXiv},
      primaryClass={cs.LG},
      url={https://arxiv.org/abs/2412.05496}, 
}

@misc{snell2024scalingllmtesttimecompute,
      title={Scaling LLM Test-Time Compute Optimally can be More Effective than Scaling Model Parameters}, 
      author={Charlie Snell and Jaehoon Lee and Kelvin Xu and Aviral Kumar},
      year={2024},
      eprint={2408.03314},
      archivePrefix={arXiv},
      primaryClass={cs.LG},
      url={https://arxiv.org/abs/2408.03314}, 
}

@misc{brown2024largelanguagemonkeysscaling,
      title={Large Language Monkeys: Scaling Inference Compute with Repeated Sampling}, 
      author={Bradley Brown and Jordan Juravsky and Ryan Ehrlich and Ronald Clark and Quoc V. Le and Christopher Ré and Azalia Mirhoseini},
      year={2024},
      eprint={2407.21787},
      archivePrefix={arXiv},
      primaryClass={cs.LG},
      url={https://arxiv.org/abs/2407.21787}, 
}

@misc{o1,
    title = {Learning to Reason with LLMs},
    url = {https://openai.com/index/learning-to-reason-with-llms/},
    author = {OpenAI},
    month = {September},
    year = {2024}
}

@misc{aime,
    title = {AIME},
    url = {https://artofproblemsolving.com/wiki/index.php/AIME_Problems_and_Solutions/},
    author = {{Mathematical Association of America}},
    month = {February},
    year = {2025}
}

@misc{huang2024olympicarenabenchmarkingmultidisciplinecognitive,
      title={OlympicArena: Benchmarking Multi-discipline Cognitive Reasoning for Superintelligent AI}, 
      author={Zhen Huang and Zengzhi Wang and Shijie Xia and Xuefeng Li and Haoyang Zou and Ruijie Xu and Run-Ze Fan and Lyumanshan Ye and Ethan Chern and Yixin Ye and Yikai Zhang and Yuqing Yang and Ting Wu and Binjie Wang and Shichao Sun and Yang Xiao and Yiyuan Li and Fan Zhou and Steffi Chern and Yiwei Qin and Yan Ma and Jiadi Su and Yixiu Liu and Yuxiang Zheng and Shaoting Zhang and Dahua Lin and Yu Qiao and Pengfei Liu},
      year={2024},
      eprint={2406.12753},
      archivePrefix={arXiv},
      primaryClass={cs.CL},
      url={https://arxiv.org/abs/2406.12753}, 
}

@misc{zhong2023agievalhumancentricbenchmarkevaluating,
      title={AGIEval: A Human-Centric Benchmark for Evaluating Foundation Models}, 
      author={Wanjun Zhong and Ruixiang Cui and Yiduo Guo and Yaobo Liang and Shuai Lu and Yanlin Wang and Amin Saied and Weizhu Chen and Nan Duan},
      year={2023},
      eprint={2304.06364},
      archivePrefix={arXiv},
      primaryClass={cs.CL},
      url={https://arxiv.org/abs/2304.06364}, 
}

@misc{ling2017programinductionrationalegeneration,
      title={Program Induction by Rationale Generation : Learning to Solve and Explain Algebraic Word Problems}, 
      author={Wang Ling and Dani Yogatama and Chris Dyer and Phil Blunsom},
      year={2017},
      eprint={1705.04146},
      archivePrefix={arXiv},
      primaryClass={cs.AI},
      url={https://arxiv.org/abs/1705.04146}, 
}

@misc{liu2020logiqachallengedatasetmachine,
      title={LogiQA: A Challenge Dataset for Machine Reading Comprehension with Logical Reasoning}, 
      author={Jian Liu and Leyang Cui and Hanmeng Liu and Dandan Huang and Yile Wang and Yue Zhang},
      year={2020},
      eprint={2007.08124},
      archivePrefix={arXiv},
      primaryClass={cs.CL},
      url={https://arxiv.org/abs/2007.08124}, 
}

@misc{gao2024omnimathuniversalolympiadlevel,
      title={Omni-MATH: A Universal Olympiad Level Mathematic Benchmark For Large Language Models}, 
      author={Bofei Gao and Feifan Song and Zhe Yang and Zefan Cai and Yibo Miao and Qingxiu Dong and Lei Li and Chenghao Ma and Liang Chen and Runxin Xu and Zhengyang Tang and Benyou Wang and Daoguang Zan and Shanghaoran Quan and Ge Zhang and Lei Sha and Yichang Zhang and Xuancheng Ren and Tianyu Liu and Baobao Chang},
      year={2024},
      eprint={2410.07985},
      archivePrefix={arXiv},
      primaryClass={cs.CL},
      url={https://arxiv.org/abs/2410.07985}, 
}

@misc{zhong2019jecqalegaldomainquestionanswering,
      title={JEC-QA: A Legal-Domain Question Answering Dataset}, 
      author={Haoxi Zhong and Chaojun Xiao and Cunchao Tu and Tianyang Zhang and Zhiyuan Liu and Maosong Sun},
      year={2019},
      eprint={1911.12011},
      archivePrefix={arXiv},
      primaryClass={cs.CL},
      url={https://arxiv.org/abs/1911.12011}, 
}

@misc{wang2021lsatprogresschallengescomplex,
      title={From LSAT: The Progress and Challenges of Complex Reasoning}, 
      author={Siyuan Wang and Zhongkun Liu and Wanjun Zhong and Ming Zhou and Zhongyu Wei and Zhumin Chen and Nan Duan},
      year={2021},
      eprint={2108.00648},
      archivePrefix={arXiv},
      primaryClass={cs.CL},
      url={https://arxiv.org/abs/2108.00648}, 
}

@misc{arora2023llmsadvancedenoughchallenging,
      title={Have LLMs Advanced Enough? A Challenging Problem Solving Benchmark For Large Language Models}, 
      author={Daman Arora and Himanshu Gaurav Singh and Mausam},
      year={2023},
      eprint={2305.15074},
      archivePrefix={arXiv},
      primaryClass={cs.CL},
      url={https://arxiv.org/abs/2305.15074}, 
}

@misc{he2024olympiadbenchchallengingbenchmarkpromoting,
      title={OlympiadBench: A Challenging Benchmark for Promoting AGI with Olympiad-Level Bilingual Multimodal Scientific Problems}, 
      author={Chaoqun He and Renjie Luo and Yuzhuo Bai and Shengding Hu and Zhen Leng Thai and Junhao Shen and Jinyi Hu and Xu Han and Yujie Huang and Yuxiang Zhang and Jie Liu and Lei Qi and Zhiyuan Liu and Maosong Sun},
      year={2024},
      eprint={2402.14008},
      archivePrefix={arXiv},
      primaryClass={cs.CL},
      url={https://arxiv.org/abs/2402.14008}, 
}

@misc{sun2024scievalmultilevellargelanguage,
      title={SciEval: A Multi-Level Large Language Model Evaluation Benchmark for Scientific Research}, 
      author={Liangtai Sun and Yang Han and Zihan Zhao and Da Ma and Zhennan Shen and Baocai Chen and Lu Chen and Kai Yu},
      year={2024},
      eprint={2308.13149},
      archivePrefix={arXiv},
      primaryClass={cs.CL},
      url={https://arxiv.org/abs/2308.13149}, 
}

@misc{chen2023theoremqatheoremdrivenquestionanswering,
      title={TheoremQA: A Theorem-driven Question Answering dataset}, 
      author={Wenhu Chen and Ming Yin and Max Ku and Pan Lu and Yixin Wan and Xueguang Ma and Jianyu Xu and Xinyi Wang and Tony Xia},
      year={2023},
      eprint={2305.12524},
      archivePrefix={arXiv},
      primaryClass={cs.CL},
      url={https://arxiv.org/abs/2305.12524}, 
}

@misc{numina_math_datasets,
  author = {Jia Li and Edward Beeching and Lewis Tunstall and Ben Lipkin and Roman Soletskyi and Shengyi Costa Huang and Kashif Rasul and Longhui Yu and Albert Jiang and Ziju Shen and Zihan Qin and Bin Dong and Li Zhou and Yann Fleureau and Guillaume Lample and Stanislas Polu},
  title = {NuminaMath},
  year = {2024},
  publisher = {Numina},
  journal = {Hugging Face repository},
  url = {https://github.com/project-numina/aimo-progress-prize/blob/main/report/numina_dataset.pdf}
}

@misc{rein2023gpqagraduatelevelgoogleproofqa,
      title={GPQA: A Graduate-Level Google-Proof Q\&A Benchmark}, 
      author={David Rein and Betty Li Hou and Asa Cooper Stickland and Jackson Petty and Richard Yuanzhe Pang and Julien Dirani and Julian Michael and Samuel R. Bowman},
      year={2023},
      eprint={2311.12022},
      archivePrefix={arXiv},
      primaryClass={cs.AI},
      url={https://arxiv.org/abs/2311.12022}, 
}

@misc{peng2022abcattentionboundedmemorycontrol,
      title={ABC: Attention with Bounded-memory Control}, 
      author={Hao Peng and Jungo Kasai and Nikolaos Pappas and Dani Yogatama and Zhaofeng Wu and Lingpeng Kong and Roy Schwartz and Noah A. Smith},
      year={2022},
      eprint={2110.02488},
      archivePrefix={arXiv},
      primaryClass={cs.CL},
      url={https://arxiv.org/abs/2110.02488}, 
}

@misc{munkhdalai2024leavecontextbehindefficient,
      title={Leave No Context Behind: Efficient Infinite Context Transformers with Infini-attention}, 
      author={Tsendsuren Munkhdalai and Manaal Faruqui and Siddharth Gopal},
      year={2024},
      eprint={2404.07143},
      archivePrefix={arXiv},
      primaryClass={cs.CL},
      url={https://arxiv.org/abs/2404.07143}, 
}

@misc{zhang2025surveytesttimescalinglarge,
      title={A Survey on Test-Time Scaling in Large Language Models: What, How, Where, and How Well?}, 
      author={Qiyuan Zhang and Fuyuan Lyu and Zexu Sun and Lei Wang and Weixu Zhang and Wenyue Hua and Haolun Wu and Zhihan Guo and Yufei Wang and Niklas Muennighoff and Irwin King and Xue Liu and Chen Ma},
      year={2025},
      eprint={2503.24235},
      archivePrefix={arXiv},
      primaryClass={cs.CL},
      url={https://arxiv.org/abs/2503.24235}, 
}

@misc{anthropic2026,
      title={Introducing Claude Opus 4.6}, 
      author={Anthropic},
      year={2026},
      url={https://www.anthropic.com/news/claude-opus-4-6}, 
}

@misc{xiao2025sliding,
  title={Why Stacking Sliding Windows Can't See Very Far},
  author={Guangxuan Xiao},
  year={2025},
  howpublished={\url{https://guangxuanx.com/blog/stacking-swa.html}}
}

@misc{xiong2021nystromformernystrombasedalgorithmapproximating,
      title={Nystr\"omformer: A Nystr\"om-Based Algorithm for Approximating Self-Attention}, 
      author={Yunyang Xiong and Zhanpeng Zeng and Rudrasis Chakraborty and Mingxing Tan and Glenn Fung and Yin Li and Vikas Singh},
      year={2021},
      eprint={2102.03902},
      archivePrefix={arXiv},
      primaryClass={cs.CL},
      url={https://arxiv.org/abs/2102.03902}, 
}

@misc{bhaskar2025cachecankvsneed,
      title={Cache Me If You Can: How Many KVs Do You Need for Effective Long-Context LMs?}, 
      author={Adithya Bhaskar and Alexander Wettig and Tianyu Gao and Yihe Dong and Danqi Chen},
      year={2025},
      eprint={2506.17121},
      archivePrefix={arXiv},
      primaryClass={cs.CL},
      url={https://arxiv.org/abs/2506.17121}, 
}

@misc{frantar2023gptqaccurateposttrainingquantization,
      title={GPTQ: Accurate Post-Training Quantization for Generative Pre-trained Transformers}, 
      author={Elias Frantar and Saleh Ashkboos and Torsten Hoefler and Dan Alistarh},
      year={2023},
      eprint={2210.17323},
      archivePrefix={arXiv},
      primaryClass={cs.LG},
      url={https://arxiv.org/abs/2210.17323}, 
}

@misc{xiao2024smoothquantaccurateefficientposttraining,
      title={SmoothQuant: Accurate and Efficient Post-Training Quantization for Large Language Models}, 
      author={Guangxuan Xiao and Ji Lin and Mickael Seznec and Hao Wu and Julien Demouth and Song Han},
      year={2024},
      eprint={2211.10438},
      archivePrefix={arXiv},
      primaryClass={cs.CL},
      url={https://arxiv.org/abs/2211.10438}, 
}

@misc{lin2024awqactivationawareweightquantization,
      title={AWQ: Activation-aware Weight Quantization for LLM Compression and Acceleration}, 
      author={Ji Lin and Jiaming Tang and Haotian Tang and Shang Yang and Wei-Ming Chen and Wei-Chen Wang and Guangxuan Xiao and Xingyu Dang and Chuang Gan and Song Han},
      year={2024},
      eprint={2306.00978},
      archivePrefix={arXiv},
      primaryClass={cs.CL},
      url={https://arxiv.org/abs/2306.00978}, 
}

@misc{akhauri2025tokenbutlertokenimportancepredictable,
      title={TokenButler: Token Importance is Predictable}, 
      author={Yash Akhauri and Ahmed F AbouElhamayed and Yifei Gao and Chi-Chih Chang and Nilesh Jain and Mohamed S. Abdelfattah},
      year={2025},
      eprint={2503.07518},
      archivePrefix={arXiv},
      primaryClass={cs.CL},
      url={https://arxiv.org/abs/2503.07518}, 
}

@misc{tang2024questqueryawaresparsityefficient,
      title={Quest: Query-Aware Sparsity for Efficient Long-Context LLM Inference}, 
      author={Jiaming Tang and Yilong Zhao and Kan Zhu and Guangxuan Xiao and Baris Kasikci and Song Han},
      year={2024},
      eprint={2406.10774},
      archivePrefix={arXiv},
      primaryClass={cs.CL},
      url={https://arxiv.org/abs/2406.10774}, 
}

@misc{lu2025mobamixtureblockattention,
      title={MoBA: Mixture of Block Attention for Long-Context LLMs}, 
      author={Enzhe Lu and Zhejun Jiang and Jingyuan Liu and Yulun Du and Tao Jiang and Chao Hong and Shaowei Liu and Weiran He and Enming Yuan and Yuzhi Wang and Zhiqi Huang and Huan Yuan and Suting Xu and Xinran Xu and Guokun Lai and Yanru Chen and Huabin Zheng and Junjie Yan and Jianlin Su and Yuxin Wu and Neo Y. Zhang and Zhilin Yang and Xinyu Zhou and Mingxing Zhang and Jiezhong Qiu},
      year={2025},
      eprint={2502.13189},
      archivePrefix={arXiv},
      primaryClass={cs.LG},
      url={https://arxiv.org/abs/2502.13189}, 
}

@misc{wang2025llmsknowdropselfattention,
      title={LLMs Know What to Drop: Self-Attention Guided KV Cache Eviction for Efficient Long-Context Inference}, 
      author={Guangtao Wang and Shubhangi Upasani and Chen Wu and Darshan Gandhi and Jonathan Li and Changran Hu and Bo Li and Urmish Thakker},
      year={2025},
      eprint={2503.08879},
      archivePrefix={arXiv},
      primaryClass={cs.CL},
      url={https://arxiv.org/abs/2503.08879}, 
}

@misc{chen2024naclgeneraleffectivekv,
      title={NACL: A General and Effective KV Cache Eviction Framework for LLMs at Inference Time}, 
      author={Yilong Chen and Guoxia Wang and Junyuan Shang and Shiyao Cui and Zhenyu Zhang and Tingwen Liu and Shuohuan Wang and Yu Sun and Dianhai Yu and Hua Wu},
      year={2024},
      eprint={2408.03675},
      archivePrefix={arXiv},
      primaryClass={cs.CL},
      url={https://arxiv.org/abs/2408.03675}, 
}

@misc{cai2025pyramidkvdynamickvcache,
      title={PyramidKV: Dynamic KV Cache Compression based on Pyramidal Information Funneling}, 
      author={Zefan Cai and Yichi Zhang and Bofei Gao and Yuliang Liu and Yucheng Li and Tianyu Liu and Keming Lu and Wayne Xiong and Yue Dong and Junjie Hu and Wen Xiao},
      year={2025},
      eprint={2406.02069},
      archivePrefix={arXiv},
      primaryClass={cs.CL},
      url={https://arxiv.org/abs/2406.02069}, 
}

@misc{li2024snapkvllmknowslooking,
      title={SnapKV: LLM Knows What You are Looking for Before Generation}, 
      author={Yuhong Li and Yingbing Huang and Bowen Yang and Bharat Venkitesh and Acyr Locatelli and Hanchen Ye and Tianle Cai and Patrick Lewis and Deming Chen},
      year={2024},
      eprint={2404.14469},
      archivePrefix={arXiv},
      primaryClass={cs.CL},
      url={https://arxiv.org/abs/2404.14469}, 
}

@misc{ge2024modeltellsdiscardadaptive,
      title={Model Tells You What to Discard: Adaptive KV Cache Compression for LLMs}, 
      author={Suyu Ge and Yunan Zhang and Liyuan Liu and Minjia Zhang and Jiawei Han and Jianfeng Gao},
      year={2024},
      eprint={2310.01801},
      archivePrefix={arXiv},
      primaryClass={cs.CL},
      url={https://arxiv.org/abs/2310.01801}, 
}

@misc{zhang2023h2oheavyhitteroracleefficient,
      title={H$_2$O: Heavy-Hitter Oracle for Efficient Generative Inference of Large Language Models}, 
      author={Zhenyu Zhang and Ying Sheng and Tianyi Zhou and Tianlong Chen and Lianmin Zheng and Ruisi Cai and Zhao Song and Yuandong Tian and Christopher Ré and Clark Barrett and Zhangyang Wang and Beidi Chen},
      year={2023},
      eprint={2306.14048},
      archivePrefix={arXiv},
      primaryClass={cs.LG},
      url={https://arxiv.org/abs/2306.14048}, 
}

@misc{jiang2024minference10acceleratingprefilling,
      title={MInference 1.0: Accelerating Pre-filling for Long-Context LLMs via Dynamic Sparse Attention}, 
      author={Huiqiang Jiang and Yucheng Li and Chengruidong Zhang and Qianhui Wu and Xufang Luo and Surin Ahn and Zhenhua Han and Amir H. Abdi and Dongsheng Li and Chin-Yew Lin and Yuqing Yang and Lili Qiu},
      year={2024},
      eprint={2407.02490},
      archivePrefix={arXiv},
      primaryClass={cs.CL},
      url={https://arxiv.org/abs/2407.02490}, 
}

@misc{fu2025mixtureattentionspansoptimizing,
      title={Mixture of Attention Spans: Optimizing LLM Inference Efficiency with Heterogeneous Sliding-Window Lengths}, 
      author={Tianyu Fu and Haofeng Huang and Xuefei Ning and Genghan Zhang and Boju Chen and Tianqi Wu and Hongyi Wang and Zixiao Huang and Shiyao Li and Shengen Yan and Guohao Dai and Huazhong Yang and Yu Wang},
      year={2025},
      eprint={2406.14909},
      archivePrefix={arXiv},
      primaryClass={cs.LG},
      url={https://arxiv.org/abs/2406.14909}, 
}

@misc{xiao2024duoattentionefficientlongcontextllm,
      title={DuoAttention: Efficient Long-Context LLM Inference with Retrieval and Streaming Heads}, 
      author={Guangxuan Xiao and Jiaming Tang and Jingwei Zuo and Junxian Guo and Shang Yang and Haotian Tang and Yao Fu and Song Han},
      year={2024},
      eprint={2410.10819},
      archivePrefix={arXiv},
      primaryClass={cs.CL},
      url={https://arxiv.org/abs/2410.10819}, 
}

@misc{chevalier2023adaptinglanguagemodelscompress,
      title={Adapting Language Models to Compress Contexts}, 
      author={Alexis Chevalier and Alexander Wettig and Anirudh Ajith and Danqi Chen},
      year={2023},
      eprint={2305.14788},
      archivePrefix={arXiv},
      primaryClass={cs.CL},
      url={https://arxiv.org/abs/2305.14788}, 
}

@misc{wang2020linformerselfattentionlinearcomplexity,
      title={Linformer: Self-Attention with Linear Complexity}, 
      author={Sinong Wang and Belinda Z. Li and Madian Khabsa and Han Fang and Hao Ma},
      year={2020},
      eprint={2006.04768},
      archivePrefix={arXiv},
      primaryClass={cs.LG},
      url={https://arxiv.org/abs/2006.04768}, 
}

@misc{child2019generatinglongsequencessparse,
      title={Generating Long Sequences with Sparse Transformers}, 
      author={Rewon Child and Scott Gray and Alec Radford and Ilya Sutskever},
      year={2019},
      eprint={1904.10509},
      archivePrefix={arXiv},
      primaryClass={cs.LG},
      url={https://arxiv.org/abs/1904.10509}, 
}

@misc{dai2019transformerxlattentivelanguagemodels,
      title={Transformer-XL: Attentive Language Models Beyond a Fixed-Length Context}, 
      author={Zihang Dai and Zhilin Yang and Yiming Yang and Jaime Carbonell and Quoc V. Le and Ruslan Salakhutdinov},
      year={2019},
      eprint={1901.02860},
      archivePrefix={arXiv},
      primaryClass={cs.LG},
      url={https://arxiv.org/abs/1901.02860}, 
}

@misc{singh2025openaigpt5card,
      title={OpenAI GPT-5 System Card}, 
      author={OpenAI},
      year={2025},
      eprint={2601.03267},
      archivePrefix={arXiv},
      primaryClass={cs.CL},
      url={https://arxiv.org/abs/2601.03267}, 
}

@misc{kazemnejad2023impactpositionalencodinglength,
      title={The Impact of Positional Encoding on Length Generalization in Transformers}, 
      author={Amirhossein Kazemnejad and Inkit Padhi and Karthikeyan Natesan Ramamurthy and Payel Das and Siva Reddy},
      year={2023},
      eprint={2305.19466},
      archivePrefix={arXiv},
      primaryClass={cs.CL},
      url={https://arxiv.org/abs/2305.19466}, 
}

@misc{yue2025doesreinforcementlearningreally,
      title={Does Reinforcement Learning Really Incentivize Reasoning Capacity in LLMs Beyond the Base Model?}, 
      author={Yang Yue and Zhiqi Chen and Rui Lu and Andrew Zhao and Zhaokai Wang and Yang Yue and Shiji Song and Gao Huang},
      year={2025},
      eprint={2504.13837},
      archivePrefix={arXiv},
      primaryClass={cs.AI},
      url={https://arxiv.org/abs/2504.13837}, 
}

@misc{openai2025gptoss120bgptoss20bmodel,
      title={gpt-oss-120b \& gpt-oss-20b Model Card}, 
      author={OpenAI},
      year={2025},
      eprint={2508.10925},
      archivePrefix={arXiv},
      primaryClass={cs.CL},
      url={https://arxiv.org/abs/2508.10925}, 
}

@misc{zhou2026sparrowsparserolloutstable,
      title={Sparrow: Sparse Rollout for Stable and Efficient Long-context RL of Large Language Models}, 
      author={Yang Zhou and Ranajoy Sadhukhan and Zhaofeng Sun and Zhuoming Chen and Souvik Kundu and Saket Dingliwal and Sai Muralidhar Jayanthi and Aram Galstyan and Haizhong Zheng and Beidi Chen},
      year={2026},
      eprint={2606.08446},
      archivePrefix={arXiv},
      primaryClass={cs.LG},
      url={https://arxiv.org/abs/2606.08446}, 
}

@misc{zheng2025actpaysefficientreinforcement,
      title={Act Only When It Pays: Efficient Reinforcement Learning for LLM Reasoning via Selective Rollouts}, 
      author={Haizhong Zheng and Yang Zhou and Brian R. Bartoldson and Bhavya Kailkhura and Fan Lai and Jiawei Zhao and Beidi Chen},
      year={2025},
      eprint={2506.02177},
      archivePrefix={arXiv},
      primaryClass={cs.AI},
      url={https://arxiv.org/abs/2506.02177}, 
}

@misc{liang2026paroquantpairwiserotationquantization,
      title={ParoQuant: Pairwise Rotation Quantization for Efficient Reasoning LLM Inference}, 
      author={Yesheng Liang and Haisheng Chen and Zihan Zhang and Song Han and Zhijian Liu},
      year={2026},
      eprint={2511.10645},
      archivePrefix={arXiv},
      primaryClass={cs.CL},
      url={https://arxiv.org/abs/2511.10645}, 
}

@misc{cai2026rkvredundancyawarekvcache,
      title={R-KV: Redundancy-aware KV Cache Compression for Reasoning Models}, 
      author={Zefan Cai and Wen Xiao and Hanshi Sun and Cheng Luo and Yikai Zhang and Ke Wan and Yucheng Li and Yeyang Zhou and Li-Wen Chang and Jiuxiang Gu and Zhen Dong and Anima Anandkumar and Abedelkadir Asi and Junjie Hu},
      year={2026},
      eprint={2505.24133},
      archivePrefix={arXiv},
      primaryClass={cs.CL},
      url={https://arxiv.org/abs/2505.24133}, 
}

@misc{dao2022flashattention,
      title={FlashAttention: Fast and Memory-Efficient Exact Attention with IO-Awareness}, 
      author={Tri Dao and Daniel Y. Fu and Stefano Ermon and Atri Rudra and Christopher Ré},
      year={2022},
      eprint={2205.14135},
      archivePrefix={arXiv},
      primaryClass={cs.LG}
}

@misc{dao2023flashattention,
      title={FlashAttention-2: Faster Attention with Better Parallelism and Work Partitioning}, 
      author={Tri Dao},
      year={2023},
      eprint={2307.08691},
      archivePrefix={arXiv},
      primaryClass={cs.LG}
}

@misc{bulatov2022recurrentmemorytransformer,
      title={Recurrent Memory Transformer}, 
      author={Aydar Bulatov and Yuri Kuratov and Mikhail S. Burtsev},
      year={2022},
      eprint={2207.06881},
      archivePrefix={arXiv},
      primaryClass={cs.CL},
      url={https://arxiv.org/abs/2207.06881}, 
}

@misc{li2025ccfcontextcompressionframework,
      title={CCF: A Context Compression Framework for Efficient Long-Sequence Language Modeling}, 
      author={Wenhao Li and Bangcheng Sun and Weihao Ye and Tianyi Zhang and Daohai Yu and Fei Chao and Rongrong Ji},
      year={2025},
      eprint={2509.09199},
      archivePrefix={arXiv},
      primaryClass={cs.CL},
      url={https://arxiv.org/abs/2509.09199}, 
}

@misc{he2025hmthierarchicalmemorytransformer,
      title={HMT: Hierarchical Memory Transformer for Efficient Long Context Language Processing}, 
      author={Zifan He and Yingqi Cao and Zongyue Qin and Neha Prakriya and Yizhou Sun and Jason Cong},
      year={2025},
      eprint={2405.06067},
      archivePrefix={arXiv},
      primaryClass={cs.CL},
      url={https://arxiv.org/abs/2405.06067}, 
}

@misc{hwang2024transformerfamfeedbackattentionworking,
      title={TransformerFAM: Feedback attention is working memory}, 
      author={Dongseong Hwang and Weiran Wang and Zhuoyuan Huo and Khe Chai Sim and Pedro Moreno Mengibar},
      year={2024},
      eprint={2404.09173},
      archivePrefix={arXiv},
      primaryClass={cs.LG},
      url={https://arxiv.org/abs/2404.09173}, 
}

@misc{ho2024blocktransformerglobaltolocallanguage,
      title={Block Transformer: Global-to-Local Language Modeling for Fast Inference}, 
      author={Namgyu Ho and Sangmin Bae and Taehyeon Kim and Hyunjik Jo and Yireun Kim and Tal Schuster and Adam Fisch and James Thorne and Se-Young Yun},
      year={2024},
      eprint={2406.02657},
      archivePrefix={arXiv},
      primaryClass={cs.CL},
      url={https://arxiv.org/abs/2406.02657}, 
}

@misc{xiang2025justthinkingefficientreasoning,
      title={Just Enough Thinking: Efficient Reasoning with Adaptive Length Penalties Reinforcement Learning}, 
      author={Violet Xiang and Chase Blagden and Rafael Rafailov and Nathan Lile and Sang Truong and Chelsea Finn and Nick Haber},
      year={2025},
      eprint={2506.05256},
      archivePrefix={arXiv},
      primaryClass={cs.AI},
      url={https://arxiv.org/abs/2506.05256}, 
}

@misc{wu2025thoughtcalibrationefficientconfident,
      title={Thought calibration: Efficient and confident test-time scaling}, 
      author={Menghua Wu and Cai Zhou and Stephen Bates and Tommi Jaakkola},
      year={2025},
      eprint={2505.18404},
      archivePrefix={arXiv},
      primaryClass={cs.LG},
      url={https://arxiv.org/abs/2505.18404}, 
}

@misc{damani2024learninghardthinkinputadaptive,
      title={Learning How Hard to Think: Input-Adaptive Allocation of LM Computation}, 
      author={Mehul Damani and Idan Shenfeld and Andi Peng and Andreea Bobu and Jacob Andreas},
      year={2024},
      eprint={2410.04707},
      archivePrefix={arXiv},
      primaryClass={cs.LG},
      url={https://arxiv.org/abs/2410.04707}, 
}

@misc{li2025steeringllmthinkingbudget,
      title={Steering LLM Thinking with Budget Guidance}, 
      author={Junyan Li and Wenshuo Zhao and Yang Zhang and Chuang Gan},
      year={2025},
      eprint={2506.13752},
      archivePrefix={arXiv},
      primaryClass={cs.CL},
      url={https://arxiv.org/abs/2506.13752}, 
}

@misc{hou2025thinkprunepruninglongchainofthought,
      title={ThinkPrune: Pruning Long Chain-of-Thought of LLMs via Reinforcement Learning}, 
      author={Bairu Hou and Yang Zhang and Jiabao Ji and Yujian Liu and Kaizhi Qian and Jacob Andreas and Shiyu Chang},
      year={2025},
      eprint={2504.01296},
      archivePrefix={arXiv},
      primaryClass={cs.CL},
      url={https://arxiv.org/abs/2504.01296}, 
}

@misc{li2025adaptivegrouppolicyoptimization,
      title={Adaptive Group Policy Optimization: Towards Stable Training and Token-Efficient Reasoning}, 
      author={Chen Li and Nazhou Liu and Kai Yang},
      year={2025},
      eprint={2503.15952},
      archivePrefix={arXiv},
      primaryClass={cs.CL},
      url={https://arxiv.org/abs/2503.15952}, 
}

@misc{shen2026dastdifficultyadaptiveslowthinkinglarge,
      title={DAST: Difficulty-Adaptive Slow-Thinking for Large Reasoning Models}, 
      author={Yi Shen and Jian Zhang and Jieyun Huang and Shuming Shi and Wenjing Zhang and Jiangze Yan and Ning Wang and Kai Wang and Zhaoxiang Liu and Shiguo Lian},
      year={2026},
      eprint={2503.04472},
      archivePrefix={arXiv},
      primaryClass={cs.LG},
      url={https://arxiv.org/abs/2503.04472}, 
}

@misc{zhao2025letlrmsbreakfree,
      title={Let LRMs Break Free from Overthinking via Self-Braking Tuning}, 
      author={Haoran Zhao and Yuchen Yan and Yongliang Shen and Haolei Xu and Wenqi Zhang and Kaitao Song and Jian Shao and Weiming Lu and Jun Xiao and Yueting Zhuang},
      year={2025},
      eprint={2505.14604},
      archivePrefix={arXiv},
      primaryClass={cs.CL},
      url={https://arxiv.org/abs/2505.14604}, 
}

@misc{dai2025sgrpoearlyexitreinforcement,
      title={S-GRPO: Early Exit via Reinforcement Learning in Reasoning Models}, 
      author={Muzhi Dai and Chenxu Yang and Qingyi Si},
      year={2025},
      eprint={2505.07686},
      archivePrefix={arXiv},
      primaryClass={cs.AI},
      url={https://arxiv.org/abs/2505.07686}, 
}

@misc{luo2025o1prunerlengthharmonizingfinetuningo1like,
      title={O1-Pruner: Length-Harmonizing Fine-Tuning for O1-Like Reasoning Pruning}, 
      author={Haotian Luo and Li Shen and Haiying He and Yibo Wang and Shiwei Liu and Wei Li and Naiqiang Tan and Xiaochun Cao and Dacheng Tao},
      year={2025},
      eprint={2501.12570},
      archivePrefix={arXiv},
      primaryClass={cs.CL},
      url={https://arxiv.org/abs/2501.12570}, 
}

@misc{liu2025understandingr1zeroliketrainingcritical,
      title={Understanding R1-Zero-Like Training: A Critical Perspective}, 
      author={Zichen Liu and Changyu Chen and Wenjun Li and Penghui Qi and Tianyu Pang and Chao Du and Wee Sun Lee and Min Lin},
      year={2025},
      eprint={2503.20783},
      archivePrefix={arXiv},
      primaryClass={cs.LG},
      url={https://arxiv.org/abs/2503.20783}, 
}

@misc{su2023roformerenhancedtransformerrotary,
      title={RoFormer: Enhanced Transformer with Rotary Position Embedding}, 
      author={Jianlin Su and Yu Lu and Shengfeng Pan and Ahmed Murtadha and Bo Wen and Yunfeng Liu},
      year={2023},
      eprint={2104.09864},
      archivePrefix={arXiv},
      primaryClass={cs.CL},
      url={https://arxiv.org/abs/2104.09864}, 
}

@article{gemini25,
  title={Gemini 2.5: Pushing the frontier with advanced reasoning, multimodality, long context, and next generation agentic capabilities},
  author={Comanici, Gheorghe and Bieber, Eric and Schaekermann, Mike and Pasupat, Ice and Sachdeva, Noveen and Dhillon, Inderjit and Blistein, Marcel and Ram, Ori and Zhang, Dan and Rosen, Evan and others},
  journal={arXiv preprint arXiv:2507.06261},
  year={2025}
}

@misc{liu2023lostmiddlelanguagemodels,
      title={Lost in the Middle: How Language Models Use Long Contexts}, 
      author={Nelson F. Liu and Kevin Lin and John Hewitt and Ashwin Paranjape and Michele Bevilacqua and Fabio Petroni and Percy Liang},
      year={2023},
      eprint={2307.03172},
      archivePrefix={arXiv},
      primaryClass={cs.CL},
      url={https://arxiv.org/abs/2307.03172}, 
}

@misc{gpt-neo,
  author       = {Black, Sid and
                  Gao, Leo and
                  Wang, Phil and
                  Leahy, Connor and
                  Biderman, Stella},
  title        = {GPT-Neo: Large Scale Autoregressive Language Modeling with Mesh-Tensorflow},
  year         = {2021},
  url          = {https://doi.org/10.5281/zenodo.5297715}
}

@misc{tandon2025endtoendtesttimetraininglong,
      title={End-to-End Test-Time Training for Long Context}, 
      author={Arnuv Tandon and Karan Dalal and Xinhao Li and Daniel Koceja and Marcel Rød and Sam Buchanan and Xiaolong Wang and Jure Leskovec and Sanmi Koyejo and Tatsunori Hashimoto and Carlos Guestrin and Jed McCaleb and Yejin Choi and Yu Sun},
      year={2025},
      eprint={2512.23675},
      archivePrefix={arXiv},
      primaryClass={cs.LG},
      url={https://arxiv.org/abs/2512.23675}, 
}

@misc{deepseekai2025deepseekv32pushingfrontieropen,
      title={DeepSeek-V3.2: Pushing the Frontier of Open Large Language Models}, 
      author={DeepSeek-AI and Aixin Liu and Aoxue Mei and Bangcai Lin and Bing Xue and Bingxuan Wang and Bingzheng Xu and Bochao Wu and Bowei Zhang and Chaofan Lin and Chen Dong and others},
      year={2025},
      eprint={2512.02556},
      archivePrefix={arXiv},
      primaryClass={cs.CL},
      url={https://arxiv.org/abs/2512.02556}, 
}

@misc{hendrycks2021measuringmathematicalproblemsolving,
      title={Measuring Mathematical Problem Solving With the MATH Dataset}, 
      author={Dan Hendrycks and Collin Burns and Saurav Kadavath and Akul Arora and Steven Basart and Eric Tang and Dawn Song and Jacob Steinhardt},
      year={2021},
      eprint={2103.03874},
      archivePrefix={arXiv},
      primaryClass={cs.LG},
      url={https://arxiv.org/abs/2103.03874}, 
}

@misc{lamb2016professorforcingnewalgorithm,
      title={Professor Forcing: A New Algorithm for Training Recurrent Networks}, 
      author={Alex Lamb and Anirudh Goyal and Ying Zhang and Saizheng Zhang and Aaron Courville and Yoshua Bengio},
      year={2016},
      eprint={1610.09038},
      archivePrefix={arXiv},
      primaryClass={stat.ML},
      url={https://arxiv.org/abs/1610.09038}, 
}

@misc{vonwerra2022trl,
  author = {Leandro von Werra and Younes Belkada and Lewis Tunstall and Edward Beeching and Tristan Thrush and Nathan Lambert and Shengyi Huang and Kashif Rasul and Quentin Gallouédec},
  title = {TRL: Transformer Reinforcement Learning},
  year = {2020},
  publisher = {GitHub},
  journal = {GitHub repository},
  howpublished = {\url{https://github.com/huggingface/trl}}
}

@misc{fu2025areallargescaleasynchronousreinforcement,
      title={AReaL: A Large-Scale Asynchronous Reinforcement Learning System for Language Reasoning}, 
      author={Wei Fu and Jiaxuan Gao and Xujie Shen and Chen Zhu and Zhiyu Mei and Chuyi He and Shusheng Xu and Guo Wei and Jun Mei and Jiashu Wang and Tongkai Yang and Binhang Yuan and Yi Wu},
      year={2025},
      eprint={2505.24298},
      archivePrefix={arXiv},
      primaryClass={cs.LG},
      url={https://arxiv.org/abs/2505.24298}, 
}

@misc{noukhovitch2025asynchronousrlhffasterefficient,
      title={Asynchronous RLHF: Faster and More Efficient Off-Policy RL for Language Models}, 
      author={Michael Noukhovitch and Shengyi Huang and Sophie Xhonneux and Arian Hosseini and Rishabh Agarwal and Aaron Courville},
      year={2025},
      eprint={2410.18252},
      archivePrefix={arXiv},
      primaryClass={cs.LG},
      url={https://arxiv.org/abs/2410.18252}, 
}

@misc{wang2024mambabytetokenfreeselectivestate,
      title={MambaByte: Token-free Selective State Space Model}, 
      author={Junxiong Wang and Tushaar Gangavarapu and Jing Nathan Yan and Alexander M. Rush},
      year={2024},
      eprint={2401.13660},
      archivePrefix={arXiv},
      primaryClass={cs.CL},
      url={https://arxiv.org/abs/2401.13660}, 
}

@misc{kitaev2020reformerefficienttransformer,
      title={Reformer: The Efficient Transformer}, 
      author={Nikita Kitaev and Łukasz Kaiser and Anselm Levskaya},
      year={2020},
      eprint={2001.04451},
      archivePrefix={arXiv},
      primaryClass={cs.LG},
      url={https://arxiv.org/abs/2001.04451}, 
}

@misc{gu2022efficientlymodelinglongsequences,
      title={Efficiently Modeling Long Sequences with Structured State Spaces}, 
      author={Albert Gu and Karan Goel and Christopher Ré},
      year={2022},
      eprint={2111.00396},
      archivePrefix={arXiv},
      primaryClass={cs.LG},
      url={https://arxiv.org/abs/2111.00396}, 
}

@misc{rabe2022selfattentiondoesneedon2,
      title={Self-attention Does Not Need $O(n^2)$ Memory}, 
      author={Markus N. Rabe and Charles Staats},
      year={2022},
      eprint={2112.05682},
      archivePrefix={arXiv},
      primaryClass={cs.LG},
      url={https://arxiv.org/abs/2112.05682}, 
}

@misc{yuan2025nativesparseattentionhardwarealigned,
      title={Native Sparse Attention: Hardware-Aligned and Natively Trainable Sparse Attention}, 
      author={Jingyang Yuan and Huazuo Gao and Damai Dai and Junyu Luo and Liang Zhao and Zhengyan Zhang and Zhenda Xie and Y. X. Wei and Lean Wang and Zhiping Xiao and Yuqing Wang and Chong Ruan and Ming Zhang and Wenfeng Liang and Wangding Zeng},
      year={2025},
      eprint={2502.11089},
      archivePrefix={arXiv},
      primaryClass={cs.CL},
      url={https://arxiv.org/abs/2502.11089}, 
}

@misc{aghajohari2025markovianthinker,
      title={The Markovian Thinker}, 
      author={Milad Aghajohari and Kamran Chitsaz and Amirhossein Kazemnejad and Sarath Chandar and Alessandro Sordoni and Aaron Courville and Siva Reddy},
      year={2025},
      eprint={2510.06557},
      archivePrefix={arXiv},
      primaryClass={cs.LG},
      url={https://arxiv.org/abs/2510.06557}, 
}

@misc{yang2024gatedlinearattentiontransformers,
      title={Gated Linear Attention Transformers with Hardware-Efficient Training}, 
      author={Songlin Yang and Bailin Wang and Yikang Shen and Rameswar Panda and Yoon Kim},
      year={2024},
      eprint={2312.06635},
      archivePrefix={arXiv},
      primaryClass={cs.LG},
      url={https://arxiv.org/abs/2312.06635}, 
}

@misc{he2025skyworkopenreasoner1,
      title={Skywork Open Reasoner 1 Technical Report}, 
      author={Jujie He and Jiacai Liu and Chris Yuhao Liu and Rui Yan and Chaojie Wang and Peng Cheng and Xiaoyu Zhang and Fuxiang Zhang and Jiacheng Xu and Wei Shen and Siyuan Li and Liang Zeng and Tianwen Wei and Cheng Cheng and Bo An and Yang Liu and Yahui Zhou},
      year={2025},
      eprint={2505.22312},
      archivePrefix={arXiv},
      primaryClass={cs.LG},
      url={https://arxiv.org/abs/2505.22312}, 
}

@misc{vaswani2017attention,
      title={Attention Is All You Need}, 
      author={Ashish Vaswani and Noam Shazeer and Niki Parmar and Jakob Uszkoreit and Llion Jones and Aidan N. Gomez and Lukasz Kaiser and Illia Polosukhin},
      year={2017},
      eprint={1706.03762},
      archivePrefix={arXiv},
      primaryClass={cs.CL}
}

@misc{peng2023rwkv,
      title={RWKV: Reinventing RNNs for the Transformer Era}, 
      author={Bo Peng and Eric Alcaide and Quentin Anthony and Alon Albalak and Samuel Arcadinho and Stella Biderman and Huanqi Cao and Xin Cheng and Michael Chung and Matteo Grella and Kranthi Kiran GV and Xuzheng He and Haowen Hou and Jiaju Lin and Przemyslaw Kazienko and Jan Kocon and Jiaming Kong and Bartlomiej Koptyra and Hayden Lau and Krishna Sri Ipsit Mantri and Ferdinand Mom and Atsushi Saito and Guangyu Song and Xiangru Tang and Bolun Wang and Johan S. Wind and Stanislaw Wozniak and Ruichong Zhang and Zhenyuan Zhang and Qihang Zhao and Peng Zhou and Qinghua Zhou and Jian Zhu and Rui-Jie Zhu},
      year={2023},
      eprint={2305.13048},
      archivePrefix={arXiv},
      primaryClass={id='cs.CL' full_name='Computation and Language' is_active=True alt_name='cmp-lg' in_archive='cs' is_general=False description='Covers natural language processing. Roughly includes material in ACM Subject Class I.2.7. Note that work on artificial languages (programming languages, logics, formal systems) that does not explicitly address natural-language issues broadly construed (natural-language processing, computational linguistics, speech, text retrieval, etc.) is not appropriate for this area.'}
}

@misc{peng2024eagle,
      title={Eagle and Finch: RWKV with Matrix-Valued States and Dynamic Recurrence}, 
      author={Bo Peng and Daniel Goldstein and Quentin Anthony and Alon Albalak and Eric Alcaide and Stella Biderman and Eugene Cheah and Xingjian Du and Teddy Ferdinan and Haowen Hou and Przemysław Kazienko and Kranthi Kiran GV and Jan Kocoń and Bartłomiej Koptyra and Satyapriya Krishna and Ronald McClelland and Jiaju Lin and Niklas Muennighoff and Fares Obeid and Atsushi Saito and Guangyu Song and Haoqin Tu and Stanisław Woźniak and Ruichong Zhang and Bingchen Zhao and Qihang Zhao and Peng Zhou and Jian Zhu and Rui-Jie Zhu},
      year={2024},
      eprint={2404.05892},
      archivePrefix={arXiv},
      primaryClass={id='cs.CL' full_name='Computation and Language' is_active=True alt_name='cmp-lg' in_archive='cs' is_general=False description='Covers natural language processing. Roughly includes material in ACM Subject Class I.2.7. Note that work on artificial languages (programming languages, logics, formal systems) that does not explicitly address natural-language issues broadly construed (natural-language processing, computational linguistics, speech, text retrieval, etc.) is not appropriate for this area.'}
}

@misc{brown2020languagemodelsfewshotlearners,
      title={Language Models are Few-Shot Learners}, 
      author={Tom B. Brown and Benjamin Mann and Nick Ryder and Melanie Subbiah and Jared Kaplan and Prafulla Dhariwal and Arvind Neelakantan and Pranav Shyam and Girish Sastry and Amanda Askell and Sandhini Agarwal and Ariel Herbert-Voss and Gretchen Krueger and Tom Henighan and Rewon Child and Aditya Ramesh and Daniel M. Ziegler and Jeffrey Wu and Clemens Winter and Christopher Hesse and Mark Chen and Eric Sigler and Mateusz Litwin and Scott Gray and Benjamin Chess and Jack Clark and Christopher Berner and Sam McCandlish and Alec Radford and Ilya Sutskever and Dario Amodei},
      year={2020},
      eprint={2005.14165},
      archivePrefix={arXiv},
      primaryClass={cs.CL},
      url={https://arxiv.org/abs/2005.14165}, 
}

@misc{beltagy2020longformer,
      title={Longformer: The Long-Document Transformer}, 
      author={Iz Beltagy and Matthew E. Peters and Arman Cohan},
      year={2020},
      eprint={2004.05150},
      archivePrefix={arXiv},
      primaryClass={cs.CL}
}

@misc{gupta2020gmatglobalmemoryaugmentation,
      title={GMAT: Global Memory Augmentation for Transformers}, 
      author={Ankit Gupta and Jonathan Berant},
      year={2020},
      eprint={2006.03274},
      archivePrefix={arXiv},
      primaryClass={cs.LG},
      url={https://arxiv.org/abs/2006.03274}, 
}

@misc{ainslie2020etcencodinglongstructured,
      title={ETC: Encoding Long and Structured Inputs in Transformers}, 
      author={Joshua Ainslie and Santiago Ontanon and Chris Alberti and Vaclav Cvicek and Zachary Fisher and Philip Pham and Anirudh Ravula and Sumit Sanghai and Qifan Wang and Li Yang},
      year={2020},
      eprint={2004.08483},
      archivePrefix={arXiv},
      primaryClass={cs.LG},
      url={https://arxiv.org/abs/2004.08483}, 
}

@misc{lu2025retrosearchexploringuntakenpaths,
      title={Retro-Search: Exploring Untaken Paths for Deeper and Efficient Reasoning}, 
      author={Ximing Lu and Seungju Han and David Acuna and Hyunwoo Kim and Jaehun Jung and Shrimai Prabhumoye and Niklas Muennighoff and Mostofa Patwary and Mohammad Shoeybi and Bryan Catanzaro and Yejin Choi},
      year={2025},
      eprint={2504.04383},
      archivePrefix={arXiv},
      primaryClass={cs.AI},
      url={https://arxiv.org/abs/2504.04383}, 
}

@misc{yong2025crosslingualreasoningtesttimescaling,
      title={Crosslingual Reasoning through Test-Time Scaling}, 
      author={Zheng-Xin Yong and M. Farid Adilazuarda and Jonibek Mansurov and Ruochen Zhang and Niklas Muennighoff and Carsten Eickhoff and Genta Indra Winata and Julia Kreutzer and Stephen H. Bach and Alham Fikri Aji},
      year={2025},
      eprint={2505.05408},
      archivePrefix={arXiv},
      primaryClass={cs.CL},
      url={https://arxiv.org/abs/2505.05408}, 
}

@misc{wang2025openhandsopenplatformai,
      title={OpenHands: An Open Platform for AI Software Developers as Generalist Agents}, 
      author={Xingyao Wang and Boxuan Li and Yufan Song and Frank F. Xu and Xiangru Tang and Mingchen Zhuge and Jiayi Pan and Yueqi Song and Bowen Li and Jaskirat Singh and Hoang H. Tran and Fuqiang Li and Ren Ma and Mingzhang Zheng and Bill Qian and Yanjun Shao and Niklas Muennighoff and Yizhe Zhang and Binyuan Hui and Junyang Lin and Robert Brennan and Hao Peng and Heng Ji and Graham Neubig},
      year={2025},
      eprint={2407.16741},
      archivePrefix={arXiv},
      primaryClass={cs.SE},
      url={https://arxiv.org/abs/2407.16741}, 
}

@misc{wang2023selfconsistencyimproveschainthought,
      title={Self-Consistency Improves Chain of Thought Reasoning in Language Models}, 
      author={Xuezhi Wang and Jason Wei and Dale Schuurmans and Quoc Le and Ed Chi and Sharan Narang and Aakanksha Chowdhery and Denny Zhou},
      year={2023},
      eprint={2203.11171},
      archivePrefix={arXiv},
      primaryClass={cs.CL},
      url={https://arxiv.org/abs/2203.11171}, 
}

@misc{ehrlich2025codemonkeysscalingtesttimecompute,
      title={CodeMonkeys: Scaling Test-Time Compute for Software Engineering}, 
      author={Ryan Ehrlich and Bradley Brown and Jordan Juravsky and Ronald Clark and Christopher Ré and Azalia Mirhoseini},
      year={2025},
      eprint={2501.14723},
      archivePrefix={arXiv},
      primaryClass={cs.LG},
      url={https://arxiv.org/abs/2501.14723}, 
}

@misc{schaeffer2025largelanguagemonkeyspower,
      title={How Do Large Language Monkeys Get Their Power (Laws)?}, 
      author={Rylan Schaeffer and Joshua Kazdan and John Hughes and Jordan Juravsky and Sara Price and Aengus Lynch and Erik Jones and Robert Kirk and Azalia Mirhoseini and Sanmi Koyejo},
      year={2025},
      eprint={2502.17578},
      archivePrefix={arXiv},
      primaryClass={cs.AI},
      url={https://arxiv.org/abs/2502.17578}, 
}

@misc{shyam2025treeattentiontopologyawaredecoding,
      title={Tree Attention: Topology-aware Decoding for Long-Context Attention on GPU clusters}, 
      author={Vasudev Shyam and Jonathan Pilault and Emily Shepperd and Quentin Anthony and Beren Millidge},
      year={2025},
      eprint={2408.04093},
      archivePrefix={arXiv},
      primaryClass={cs.LG},
      url={https://arxiv.org/abs/2408.04093}, 
}

@misc{liu2023ringattentionblockwisetransformers,
      title={Ring Attention with Blockwise Transformers for Near-Infinite Context}, 
      author={Hao Liu and Matei Zaharia and Pieter Abbeel},
      year={2023},
      eprint={2310.01889},
      archivePrefix={arXiv},
      primaryClass={cs.CL},
      url={https://arxiv.org/abs/2310.01889}, 
}

@misc{dao2024transformersssmsgeneralizedmodels,
      title={Transformers are SSMs: Generalized Models and Efficient Algorithms Through Structured State Space Duality}, 
      author={Tri Dao and Albert Gu},
      year={2024},
      eprint={2405.21060},
      archivePrefix={arXiv},
      primaryClass={cs.LG},
      url={https://arxiv.org/abs/2405.21060}, 
}

@misc{sadhukhan2025kineticsrethinkingtesttimescaling,
      title={Kinetics: Rethinking Test-Time Scaling Laws}, 
      author={Ranajoy Sadhukhan and Zhuoming Chen and Haizhong Zheng and Yang Zhou and Emma Strubell and Beidi Chen},
      year={2025},
      eprint={2506.05333},
      archivePrefix={arXiv},
      primaryClass={cs.LG},
      url={https://arxiv.org/abs/2506.05333}, 
}

@misc{wu2026rlboostharvestingpreemptibleresources,
      title={RLBoost: Harvesting Preemptible Resources for Cost-Efficient Reinforcement Learning on LLMs}, 
      author={Yongji Wu and Xueshen Liu and Haizhong Zheng and Juncheng Gu and Beidi Chen and Z. Morley Mao and Arvind Krishnamurthy and Ion Stoica},
      year={2026},
      eprint={2510.19225},
      archivePrefix={arXiv},
      primaryClass={cs.DC},
      url={https://arxiv.org/abs/2510.19225}, 
}

@misc{dong2026scalablellmreasoningacceleration,
      title={Scalable LLM Reasoning Acceleration with Low-rank Distillation}, 
      author={Harry Dong and Bilge Acun and Beidi Chen and Yuejie Chi},
      year={2026},
      eprint={2505.07861},
      archivePrefix={arXiv},
      primaryClass={cs.CL},
      url={https://arxiv.org/abs/2505.07861}, 
}

@misc{wu2025resumunlockinglonghorizonsearch,
      title={ReSum: Unlocking Long-Horizon Search Intelligence via Context Summarization}, 
      author={Xixi Wu and Kuan Li and Yida Zhao and Liwen Zhang and Litu Ou and Huifeng Yin and Zhongwang Zhang and Yong Jiang and Pengjun Xie and Fei Huang and Minhao Cheng and Shuai Wang and Hong Cheng and Jingren Zhou},
      year={2025},
      eprint={2509.13313},
      archivePrefix={arXiv},
      primaryClass={cs.CL},
      url={https://arxiv.org/abs/2509.13313}, 
}

@article{tay2020long,
  title={Long range arena: A benchmark for efficient transformers},
  author={Tay, Yi and Dehghani, Mostafa and Abnar, Samira and Shen, Yikang and Bahri, Dara and Pham, Philip and Rao, Jinfeng and Yang, Liu and Ruder, Sebastian and Metzler, Donald},
  journal={arXiv preprint arXiv:2011.04006},
  year={2020}
}

@misc{jaegle2021perceiver,
      title={Perceiver: General Perception with Iterative Attention}, 
      author={Andrew Jaegle and Felix Gimeno and Andrew Brock and Andrew Zisserman and Oriol Vinyals and Joao Carreira},
      year={2021},
      eprint={2103.03206},
      archivePrefix={arXiv},
      primaryClass={cs.CV}
}

@misc{shao2024deepseekmathpushinglimitsmathematical,
      title={DeepSeekMath: Pushing the Limits of Mathematical Reasoning in Open Language Models}, 
      author={Zhihong Shao and Peiyi Wang and Qihao Zhu and Runxin Xu and Junxiao Song and Xiao Bi and Haowei Zhang and Mingchuan Zhang and Y. K. Li and Y. Wu and Daya Guo},
      year={2024},
      eprint={2402.03300},
      archivePrefix={arXiv},
      primaryClass={cs.CL},
      url={https://arxiv.org/abs/2402.03300}, 
}

@inproceedings{kwon2023efficient,
  title={Efficient Memory Management for Large Language Model Serving with PagedAttention},
  author={Woosuk Kwon and Zhuohan Li and Siyuan Zhuang and Ying Sheng and Lianmin Zheng and Cody Hao Yu and Joseph E. Gonzalez and Hao Zhang and Ion Stoica},
  booktitle={Proceedings of the ACM SIGOPS 29th Symposium on Operating Systems Principles},
  year={2023}
}

@misc{liu2023scissorhandsexploitingpersistenceimportance,
      title={Scissorhands: Exploiting the Persistence of Importance Hypothesis for LLM KV Cache Compression at Test Time}, 
      author={Zichang Liu and Aditya Desai and Fangshuo Liao and Weitao Wang and Victor Xie and Zhaozhuo Xu and Anastasios Kyrillidis and Anshumali Shrivastava},
      year={2023},
      eprint={2305.17118},
      archivePrefix={arXiv},
      primaryClass={cs.LG},
      url={https://arxiv.org/abs/2305.17118}, 
}

@misc{chang2026valueawarestochastickvcache,
      title={Value-Aware Stochastic KV Cache Eviction for Reasoning Models}, 
      author={Ting-Yun Chang and Harvey Yiyun Fu and Deqing Fu and Chenghao Yang and Jesse Thomason and Robin Jia},
      year={2026},
      eprint={2606.03928},
      archivePrefix={arXiv},
      primaryClass={cs.LG},
      url={https://arxiv.org/abs/2606.03928}, 
}

@misc{wang2026crystalkvefficientkvcache,
      title={Crystal-KV: Efficient KV Cache Management for Chain-of-Thought LLMs via Answer-First Principle}, 
      author={Zihan Wang and Cheng Tang and Lei Gong and Cheng Li and Chao Wang and Teng Wang and Wenqi Lou and Xuehai Zhou},
      year={2026},
      eprint={2601.16986},
      archivePrefix={arXiv},
      primaryClass={cs.CL},
      url={https://arxiv.org/abs/2601.16986}, 
}

@misc{ramachandran2026thinkvthoughtadaptivekvcache,
      title={ThinKV: Thought-Adaptive KV Cache Compression for Efficient Reasoning Models}, 
      author={Akshat Ramachandran and Marina Neseem and Charbel Sakr and Rangharajan Venkatesan and Brucek Khailany and Tushar Krishna},
      year={2026},
      eprint={2510.01290},
      archivePrefix={arXiv},
      primaryClass={cs.LG},
      url={https://arxiv.org/abs/2510.01290}, 
}

@misc{metel2026thinkinglongshortstable,
      title={Thinking Long, but Short: Stable Sequential Test-Time Scaling for Large Reasoning Models}, 
      author={Michael R. Metel and Yufei Cui and Boxing Chen and Prasanna Parthasarathi},
      year={2026},
      eprint={2601.09855},
      archivePrefix={arXiv},
      primaryClass={cs.AI},
      url={https://arxiv.org/abs/2601.09855}, 
}

@misc{zhang2026recursivelanguagemodels,
      title={Recursive Language Models}, 
      author={Alex L. Zhang and Tim Kraska and Omar Khattab},
      year={2026},
      eprint={2512.24601},
      archivePrefix={arXiv},
      primaryClass={cs.AI},
      url={https://arxiv.org/abs/2512.24601}, 
}

@misc{han2026karaefficientreasoningllm,
      title={KARA: Efficient Reasoning LLM Serving via Sliding-Window KV Cache Compression}, 
      author={Shen Han and Yuyang Wu and Junpu Yu and Olexandr Isayev},
      year={2026},
      eprint={2607.01237},
      archivePrefix={arXiv},
      primaryClass={cs.CL},
      url={https://arxiv.org/abs/2607.01237}, 
}

@inproceedings{vajipey2025simple,
  title={Simple, Scalable Reasoning via Iterated Summarization},
  author={Vajipey, Vivek and Tadimeti, Aditya and Shen, Justin and Prystawski, Ben and Li, Michael Y and Goodman, Noah},
  year={2025},
  url={https://openreview.net/pdf?id=uhZLKclfGB}
}

@misc{yang2025qwen3technicalreport,
      title={Qwen3 Technical Report}, 
      author={An Yang and Anfeng Li and Baosong Yang and Beichen Zhang and Binyuan Hui and Bo Zheng and Bowen Yu and Chang Gao and Chengen Huang and Chenxu Lv and others},
      year={2025},
      eprint={2505.09388},
      archivePrefix={arXiv},
      primaryClass={cs.CL},
      url={https://arxiv.org/abs/2505.09388}, 
}

@misc{yu2026swaaslidingwindowattention,
      title={SWAA: Sliding Window Attention Adaptation for Efficient and Quality Preserving Long Context Processing}, 
      author={Yijiong Yu and Jiale Liu and Qingyun Wu and Huazheng Wang and Ji Pei},
      year={2026},
      eprint={2512.10411},
      archivePrefix={arXiv},
      primaryClass={cs.CL},
      url={https://arxiv.org/abs/2512.10411}, 
}

@misc{lanham2023measuringfaithfulnesschainofthoughtreasoning,
      title={Measuring Faithfulness in Chain-of-Thought Reasoning}, 
      author={Tamera Lanham and Anna Chen and Ansh Radhakrishnan and Benoit Steiner and Carson Denison and Danny Hernandez and Dustin Li and Esin Durmus and Evan Hubinger and Jackson Kernion and Kamilė Lukošiūtė and Karina Nguyen and Newton Cheng and Nicholas Joseph and Nicholas Schiefer and Oliver Rausch and Robin Larson and Sam McCandlish and Sandipan Kundu and Saurav Kadavath and Shannon Yang and Thomas Henighan and Timothy Maxwell and Timothy Telleen-Lawton and Tristan Hume and Zac Hatfield-Dodds and Jared Kaplan and Jan Brauner and Samuel R. Bowman and Ethan Perez},
      year={2023},
      eprint={2307.13702},
      archivePrefix={arXiv},
      primaryClass={cs.AI},
      url={https://arxiv.org/abs/2307.13702}, 
}

@misc{guha2025openthoughtsdatarecipesreasoning,
      title={OpenThoughts: Data Recipes for Reasoning Models}, 
      author={Etash Guha and Ryan Marten and Sedrick Keh and Negin Raoof and Georgios Smyrnis and Hritik Bansal and Marianna Nezhurina and Jean Mercat and Trung Vu and Zayne Sprague and Ashima Suvarna and Benjamin Feuer and Liangyu Chen and Zaid Khan and Eric Frankel and Sachin Grover and Caroline Choi and Niklas Muennighoff and Shiye Su and Wanjia Zhao and John Yang and Shreyas Pimpalgaonkar and Kartik Sharma and Charlie Cheng-Jie Ji and Yichuan Deng and Sarah Pratt and Vivek Ramanujan and Jon Saad-Falcon and Jeffrey Li and Achal Dave and Alon Albalak and Kushal Arora and Blake Wulfe and Chinmay Hegde and Greg Durrett and Sewoong Oh and Mohit Bansal and Saadia Gabriel and Aditya Grover and Kai-Wei Chang and Vaishaal Shankar and Aaron Gokaslan and Mike A. Merrill and Tatsunori Hashimoto and Yejin Choi and Jenia Jitsev and Reinhard Heckel and Maheswaran Sathiamoorthy and Alexandros G. Dimakis and Ludwig Schmidt},
      year={2025},
      eprint={2506.04178},
      archivePrefix={arXiv},
      primaryClass={cs.LG},
      url={https://arxiv.org/abs/2506.04178}, 
}

@misc{zaheer2021bigbirdtransformerslonger,
      title={Big Bird: Transformers for Longer Sequences}, 
      author={Manzil Zaheer and Guru Guruganesh and Avinava Dubey and Joshua Ainslie and Chris Alberti and Santiago Ontanon and Philip Pham and Anirudh Ravula and Qifan Wang and Li Yang and Amr Ahmed},
      year={2021},
      eprint={2007.14062},
      archivePrefix={arXiv},
      primaryClass={cs.LG},
      url={https://arxiv.org/abs/2007.14062}, 
}

@misc{gu2024mambalineartimesequencemodeling,
      title={Mamba: Linear-Time Sequence Modeling with Selective State Spaces}, 
      author={Albert Gu and Tri Dao},
      year={2024},
      eprint={2312.00752},
      archivePrefix={arXiv},
      primaryClass={cs.LG},
      url={https://arxiv.org/abs/2312.00752}, 
}

@misc{aggarwal2025l1controllinglongreasoning,
      title={L1: Controlling How Long A Reasoning Model Thinks With Reinforcement Learning}, 
      author={Pranjal Aggarwal and Sean Welleck},
      year={2025},
      eprint={2503.04697},
      archivePrefix={arXiv},
      primaryClass={cs.CL},
      url={https://arxiv.org/abs/2503.04697}, 
}

@misc{xiao2024efficientstreaminglanguagemodels,
      title={Efficient Streaming Language Models with Attention Sinks}, 
      author={Guangxuan Xiao and Yuandong Tian and Beidi Chen and Song Han and Mike Lewis},
      year={2024},
      eprint={2309.17453},
      archivePrefix={arXiv},
      primaryClass={cs.CL},
      url={https://arxiv.org/abs/2309.17453}, 
}

@misc{fu2025slidingwindowattentiontraining,
      title={Sliding Window Attention Training for Efficient Large Language Models}, 
      author={Zichuan Fu and Wentao Song and Yejing Wang and Xian Wu and Yefeng Zheng and Yingying Zhang and Derong Xu and Xuetao Wei and Tong Xu and Xiangyu Zhao},
      year={2025},
      eprint={2502.18845},
      archivePrefix={arXiv},
      primaryClass={cs.CL},
      url={https://arxiv.org/abs/2502.18845}, 
}

@misc{hu2025raasreasoningawareattentionsparsity,
      title={RaaS: Reasoning-Aware Attention Sparsity for Efficient LLM Reasoning}, 
      author={Junhao Hu and Wenrui Huang and Weidong Wang and Zhenwen Li and Tiancheng Hu and Zhixia Liu and Xusheng Chen and Tao Xie and Yizhou Shan},
      year={2025},
      eprint={2502.11147},
      archivePrefix={arXiv},
      primaryClass={cs.LG},
      url={https://arxiv.org/abs/2502.11147}, 
}

@misc{zhang2025lighttransferlongcontextllmsecretly,
      title={LightTransfer: Your Long-Context LLM is Secretly a Hybrid Model with Effortless Adaptation}, 
      author={Xuan Zhang and Fengzhuo Zhang and Cunxiao Du and Chao Du and Tianyu Pang and Wei Gao and Min Lin},
      year={2025},
      eprint={2410.13846},
      archivePrefix={arXiv},
      primaryClass={cs.CL},
      url={https://arxiv.org/abs/2410.13846}, 
}

@misc{rae2019compressivetransformerslongrangesequence,
      title={Compressive Transformers for Long-Range Sequence Modelling}, 
      author={Jack W. Rae and Anna Potapenko and Siddhant M. Jayakumar and Timothy P. Lillicrap},
      year={2019},
      eprint={1911.05507},
      archivePrefix={arXiv},
      primaryClass={cs.LG},
      url={https://arxiv.org/abs/1911.05507}, 
}

@misc{gema2025inversescalingtesttimecompute,
      title={Inverse Scaling in Test-Time Compute}, 
      author={Aryo Pradipta Gema and Alexander Hägele and Runjin Chen and Andy Arditi and Jacob Goldman-Wetzler and Kit Fraser-Taliente and Henry Sleight and Linda Petrini and Julian Michael and Beatrice Alex and Pasquale Minervini and Yanda Chen and Joe Benton and Ethan Perez},
      year={2025},
      eprint={2507.14417},
      archivePrefix={arXiv},
      primaryClass={cs.AI},
      url={https://arxiv.org/abs/2507.14417}, 
}

@misc{wu2021recursivelysummarizingbookshuman,
      title={Recursively Summarizing Books with Human Feedback}, 
      author={Jeff Wu and Long Ouyang and Daniel M. Ziegler and Nisan Stiennon and Ryan Lowe and Jan Leike and Paul Christiano},
      year={2021},
      eprint={2109.10862},
      archivePrefix={arXiv},
      primaryClass={cs.CL},
      url={https://arxiv.org/abs/2109.10862}, 
}

@misc{choromanski2022rethinkingattentionperformers,
      title={Rethinking Attention with Performers}, 
      author={Krzysztof Choromanski and Valerii Likhosherstov and David Dohan and Xingyou Song and Andreea Gane and Tamas Sarlos and Peter Hawkins and Jared Davis and Afroz Mohiuddin and Lukasz Kaiser and David Belanger and Lucy Colwell and Adrian Weller},
      year={2022},
      eprint={2009.14794},
      archivePrefix={arXiv},
      primaryClass={cs.LG},
      url={https://arxiv.org/abs/2009.14794}, 
}

@misc{muennighoff2025s1simpletesttimescaling,
      title={s1: Simple test-time scaling}, 
      author={Niklas Muennighoff and Zitong Yang and Weijia Shi and Xiang Lisa Li and Li Fei-Fei and Hannaneh Hajishirzi and Luke Zettlemoyer and Percy Liang and Emmanuel Candès and Tatsunori Hashimoto},
      year={2025},
      eprint={2501.19393},
      archivePrefix={arXiv},
      primaryClass={cs.CL},
      url={https://arxiv.org/abs/2501.19393}, 
}

@misc{yan2025inftythinkbreakinglengthlimits,
      title={InftyThink: Breaking the Length Limits of Long-Context Reasoning in Large Language Models}, 
      author={Yuchen Yan and Yongliang Shen and Yang Liu and Jin Jiang and Mengdi Zhang and Jian Shao and Yueting Zhuang},
      year={2025},
      eprint={2503.06692},
      archivePrefix={arXiv},
      primaryClass={cs.CL},
      url={https://arxiv.org/abs/2503.06692}, 
}
\bibliographystyle{colm2026_conference}

\clearpage

\appendix
\section{Extended Related Work}
\label{sec:extrel}

\paragraph{Key-Value (KV) Cache} The KV cache in transformers stores information from the past context, which eventually grows prohibitively expensive as generation continues. Thus, many context extensions focus specifically on handling the KV cache footprint. Methods either target the KV-cache from pre-fill, post-fill, or both~\citep{bhaskar2025cachecankvsneed}. Pre-fill methods seek to reduce the cost of the KV-cache from the prompt~\citep{jiang2024minference10acceleratingprefilling,eyuboglu2025cartridgeslightweightgeneralpurposelong}, while post-fill methods deal with the KV cache after processing the prompt~\citep{zhang2023h2oheavyhitteroracleefficient,liu2023scissorhandsexploitingpersistenceimportance,li2024snapkvllmknowslooking,ge2024modeltellsdiscardadaptive,chen2024naclgeneraleffectivekv,wang2025llmsknowdropselfattention,cai2025pyramidkvdynamickvcache,hu2025raasreasoningawareattentionsparsity,cai2026rkvredundancyawarekvcache,metel2026thinkinglongshortstable,ramachandran2026thinkvthoughtadaptivekvcache,wang2026crystalkvefficientkvcache,chang2026valueawarestochastickvcache,zweiger2026fastkvcompactionattention}. Many post-fill methods use recency eviction methods to discard older parts of the KV cache~\citep{xiao2024efficientstreaminglanguagemodels,xiao2024duoattentionefficientlongcontextllm,fu2025mixtureattentionspansoptimizing,han2026karaefficientreasoningllm,yu2026swaaslidingwindowattention}. Other approaches across pre-fill and post-fill optimize the KV cache and memory usage by taking hardware into consideration~\citep{tang2024questqueryawaresparsityefficient,lu2025mobamixtureblockattention,akhauri2025tokenbutlertokenimportancepredictable,yuan2025nativesparseattentionhardwarealigned} or using quantization techniques~\citep{lin2024awqactivationawareweightquantization,xiao2024smoothquantaccurateefficientposttraining,frantar2023gptqaccurateposttrainingquantization}. Importantly, \method does not reduce the pre-fill cost of the KV cache, which may lead to high memory usage with extremely long prefixes. One solution could be to discard information from the prompt that is not needed for the prefix, or to introduce context management techniques~\citep{zhang2026recursivelanguagemodels}. Long inputs, such as a relevant book to solve the problem, are likely better suited in a file whose path is provided to the model so it can read it step by step.

\paragraph{Efficient thinking} Several works have explored improving the thinking efficiency of reasoning language models. They seek to do so at train-time, test-time, or both. Train-time methods often target the RL algorithm or infrastructure-related issues~\citep{liu2025understandingr1zeroliketrainingcritical,luo2025o1prunerlengthharmonizingfinetuningo1like,dai2025sgrpoearlyexitreinforcement,zhao2025letlrmsbreakfree,shen2026dastdifficultyadaptiveslowthinkinglarge,li2025adaptivegrouppolicyoptimization,hou2025thinkprunepruninglongchainofthought,xiang2025justthinkingefficientreasoning,zheng2025actpaysefficientreinforcement,zhou2026sparrowsparserolloutstable,wu2026rlboostharvestingpreemptibleresources}, while test-time methods seek to work with existing models out-of-the-box that have already been trained~\citep{damani2024learninghardthinkinputadaptive,li2025steeringllmthinkingbudget,wu2025thoughtcalibrationefficientconfident,liang2026paroquantpairwiserotationquantization,dong2026scalablellmreasoningacceleration,muennighoff2025s1simpletesttimescaling}. \method is both a train-time and test-time method: It can be used during RL and can be applied to already-trained models to achieve better efficiency and performance. 

\section{Other sliding window sizes}
\label{sec:othersw}

\begin{figure}[h]
\centering
\includegraphics[width=\columnwidth]{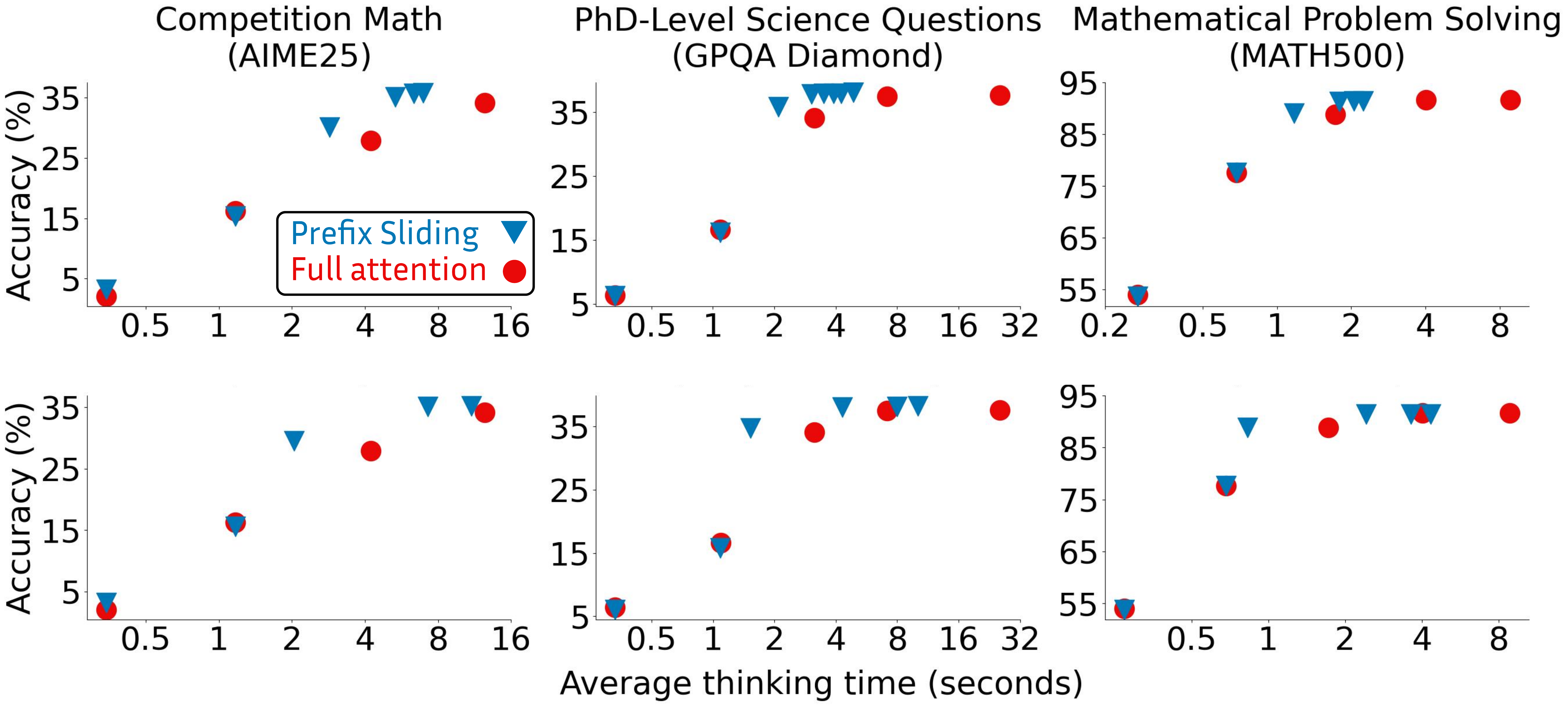}
\caption{\textbf{\autoref{fig:1} with sliding window size 8192 (top) and 16384 (bottom).}}
\label{fig:13}
\end{figure}

\FloatBarrier
\clearpage

\section{Tabular results}
\label{sec:tab}

\begin{table}[htbp]
\centering
\begin{tabular}{l|rrrrrr|rr}
\toprule
Window & \multicolumn{2}{c}{\textbf{AIME25}} & \multicolumn{2}{c}{\textbf{GPQA}} & \multicolumn{2}{c|}{\textbf{MATH500}} & \multicolumn{2}{c}{\textbf{Tok/s at}} \\
size & avg@64 & avglen & avg@64 & avglen & avg@64 & avglen & 32K & 128K \\
\midrule
2048 & 27.7 & 47643 & 35.9 & 30107 & 89.8 & 9310 & 8973 & 8737 \\
4096 & 33.9 & 29943 & 37.0 & 16707 & 91.5 & 7069 & 5479 & 5224 \\
8192 & 35.8 & 19373 & 38.0 & 13605 & 91.4 & 6229 & 3291 & 2788 \\
16384 & 35.3 & 19872 & 38.2 & 14378 & 91.5 & 6160 & 2441 & 1420 \\
Full & 34.2 & 19158 & 37.6 & 11403 & 91.7 & 6056 & 1477 & 448 \\
\bottomrule
\end{tabular}
\caption{\textbf{\method achieves performance comparable to full attention at significantly higher speeds}. These performance numbers are used for figures throughout the paper.}
\label{tab:sw}
\end{table}

\FloatBarrier

\section{Continue vs Reset PE}
\label{sec:pe}

\begin{figure}[h]
\centering
\includegraphics[width=\columnwidth]{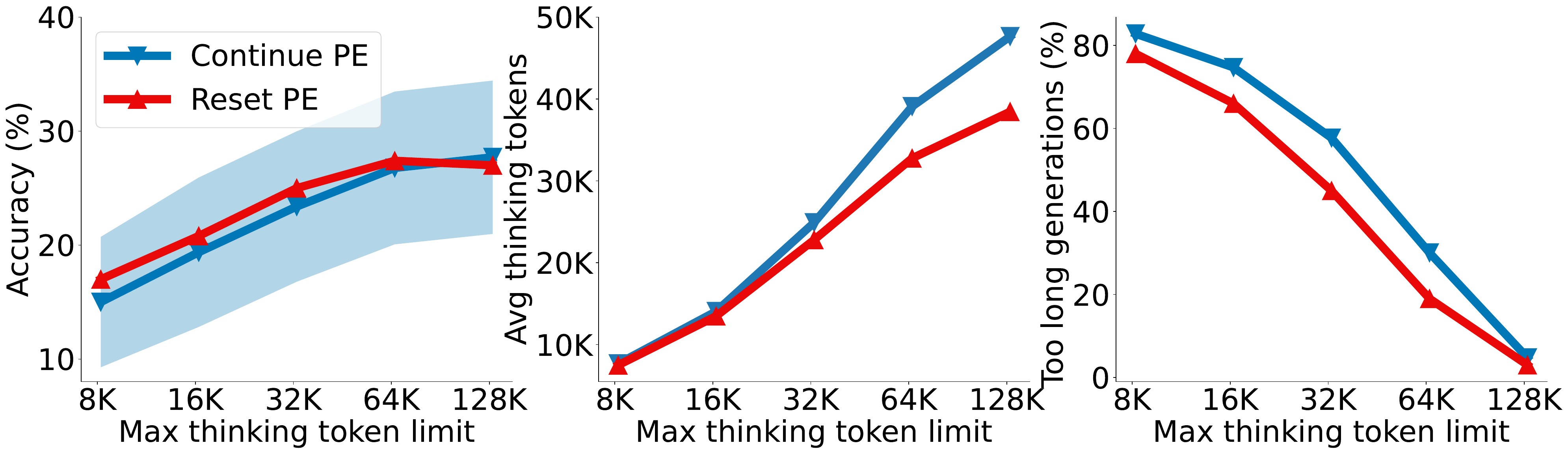}
\caption{\textbf{Continue PE performs similarly to reset PE.} The benchmark is AIME25 and the shaded area represents the standard error. The sliding window size is 2048.}
\label{fig:14}
\end{figure}

\section{Truncated backpropagation validation}
\label{sec:trunc}

\begin{figure}[h]
\centering
\includegraphics[width=\columnwidth]{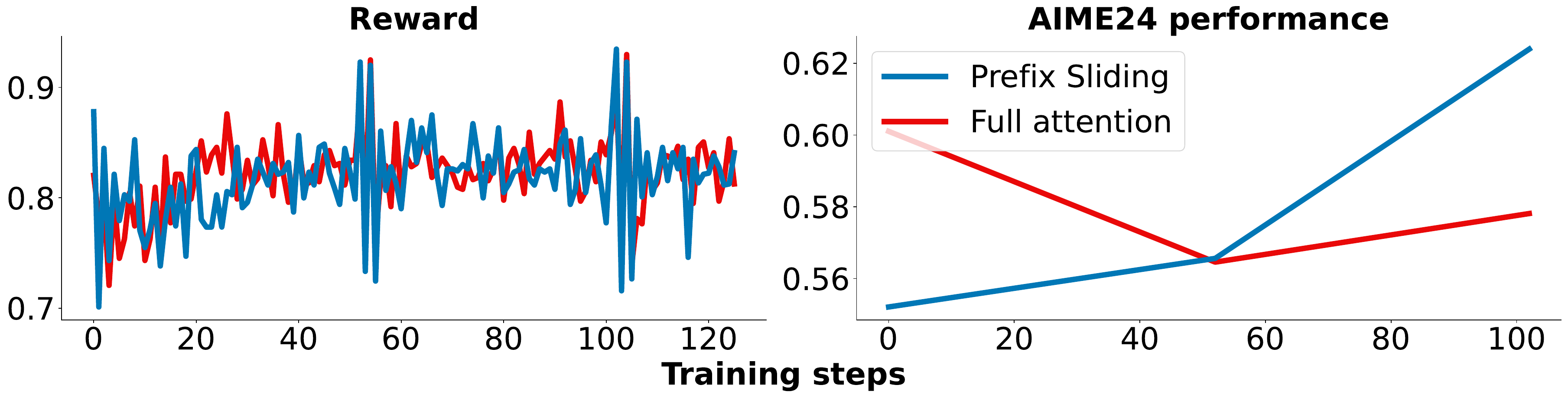}
\caption{\textbf{\method with truncated backpropagation can match full attention.}}
\label{fig:15}
\end{figure}

\autoref{fig:15} depicts a short experiment with DeepSeek-R1-Distill-Qwen-7B~\citep{r1} using the prime-rl codebase for asynchronous reinforcement learning~\citep{primeintellect2025prime-rl}. For both models, 32768 tokens are passed to the trainer, but for \method only 8,192 of them are backpropagated using truncated backpropagation (\autoref{sec:method}) with a multiplier of 4. The window size for \method is 8,192. We find that performance is comparable, but highlight that more experiments at even larger scales are necessary in the future. Therefore, we stick with truncated backpropagation for our experiments (e.g., \autoref{fig:7}), but larger-scale runs with significantly longer chains may require chunked backpropagation (\autoref{sec:method}).

\section{Training Dataset}
\label{sec:data}

For our dataset for reinforcement learning training we combine public sources, specifically SkyWork~\citep{he2025skyworkopenreasoner1} and s1~\citep{muennighoff2025s1simpletesttimescaling} (s1 further sources from NuminaMATH~\citep{numina_math_datasets}, MATH~\citep{hendrycks2021measuringmathematicalproblemsolving}, OlympicArena~\citep{huang2024olympicarenabenchmarkingmultidisciplinecognitive}, OmniMath~\citep{gao2024omnimathuniversalolympiadlevel}, AGIEval~\citep{zhong2023agievalhumancentricbenchmarkevaluating,ling2017programinductionrationalegeneration,hendrycks2021measuringmathematicalproblemsolving,liu2020logiqachallengedatasetmachine,zhong2019jecqalegaldomainquestionanswering,wang2021lsatprogresschallengescomplex}, OlympiadBench~\citep{he2024olympiadbenchchallengingbenchmarkpromoting}, TheoremQA~\citep{chen2023theoremqatheoremdrivenquestionanswering}, JEEBench~\citep{arora2023llmsadvancedenoughchallenging}, GPQA~\citep{rein2023gpqagraduatelevelgoogleproofqa}, SciEval~\citep{sun2024scievalmultilevellargelanguage}). We also decontaminate against test data using the s1 setup.

To filter problems, we rely on three criteria: guessability, verifiability, and difficulty. For guessability, we remove samples where small models write the correct solution on any of 8 tries without thinking~\citep{k1.5}. For verifiability, we remove any samples that contain a set of words such as ``How'', ``Explain'', as such questions may have answers that cannot be easily objectively verified. For difficulty, we score models on all samples multiple times. We then remove those always solved by weak models, as well as those never solved by strong models, as they may be impossible to solve.

\section{Other methods}
\label{sec:othermethods}

\FloatBarrier

\subsection{Last k hyperparameters}
\label{sec:lastk}

\begin{wraptable}{r}{0.43\textwidth}
\centering
\begin{tabular}{l|rr}
\toprule
\textbf{$k$ tokens} & \textbf{MATH500} & \textbf{AIME25} \\
\midrule
64   & 58.2 & 3.2 \\
128  & 59.7 & 3.5 \\
256  & \textbf{60.8} & \textbf{4.2} \\
512  & 60.3 & \textbf{4.2} \\
1024 & 54.6 & 2.4 \\
\bottomrule
\end{tabular}
\caption{\textbf{Ablating last $k$.} Results are avg@64 with a 2048-token context and one pass. We select $k=256$, which performs best while keeping the carried context small.}
\label{tab:ablatek}
\end{wraptable}

We select $k=256$ for our runs in \autoref{fig:9} based on a sweep in \autoref{tab:ablatek}. This means that whenever the model runs out of context, exactly 256 tokens from the end of the generation are taken and prepended to the thinking in the next generation of the model. We do not consider proper sentence endings; thus, the tokens are likely to be cut off; see \autoref{fig:lastkissue} for an example. To keep the total thinking tokens the same after the model has received the last k tokens of the first turn, it generates $k$ fewer tokens from the second turn onward. The optimal value of last k likely depends on the context window and the number of allowed passes, so the setup in \autoref{fig:9} may have a different optimum. However, in practice one cannot predict the number of passes ahead of time, but generally needs to use the same last k value across setups.

\subsection{Summary hyperparameters}
\label{sec:summ}

We set a maximum summary length of $k=256$ based on \autoref{sec:lastk}. We treat the summary as a tool call and use summary forcing: If $k$ tokens are left in the context window, and the summary tool has not yet been invoked, we force-insert the tool call into the reasoning of the model so it generates a summary. We use the model itself as the summary model. We ablate prompts in \autoref{tab:prompt}. Adding an example of how to use the tool and context (prompt 2) raises performance; possibly it leads to better summaries and their usage. Explaining the context further (prompt 3) did not help. \autoref{fig:summissue} shows an issue with the summary approach where the model seemingly ignores its summary, or maybe it uses it without mentioning it in its thinking~\citep{lanham2023measuringfaithfulnesschainofthoughtreasoning,chen2025reasoningmodelsdontsay}. Thus, we stick with prompt 2 for \autoref{fig:9}. For summary, we do not subtract the summary length from the tokens the model may generate in its second turn onward; thus, it keeps slightly more tokens in memory than \method or last k. This may give the summary approach a slight advantage in \autoref{fig:9}. We also tried inserting the summary in the model's thinking, which led to worse performance; possibly it was very out-of-distribution for the model.

\begin{table}
\centering
\begin{tabular}{l|rr}
\toprule
\textbf{Prompt} & \multicolumn{2}{c}{\textbf{AIME25}} \\
& \textbf{Accuracy} & \textbf{Coverage} \\
\midrule
~~~1: Tool only (\autoref{fig:promptsystem}, \autoref{fig:promptusercontext}) & 23.2 & 33.3 \\
~~~2: Tool/context examples (\autoref{fig:promptsystemshot}, \autoref{fig:promptusercontext}) & 26.4 & 53.3 \\
~~~3: Tool/context examples with info (\autoref{fig:promptsystemshot}, \autoref{fig:promptuserwarning}) & 25.8 & 46.7 \\
\bottomrule
\end{tabular}
\caption{\textbf{Summary prompt ablations.} The first figure in the brackets is the system prompt; the second is the user prompt after a summary is produced (the first turn is unmodified). cov@64 is coverage across 64 samples, i.e., if any of the 64 are correct, it is counted as correct.}
\label{tab:prompt}
\end{table}

\begin{figure}
\centering
\displaycode{
\# Tools

~

You may call one or more functions to assist with the user query.

~

You are provided with function signatures within <tools></tools> XML tags:

<tools>

\{"type": "function", "function": \{"name": "pass", "description": "Passes the task to the next model to solve the problem from where you left off. Useful when thinking gets too long.", "parameters": \{"type": "object", "properties": \{"context": \{"type": "string", "description": "Information to pass to the next model. E.g., ideas already tried, key results, next steps to perform..."\}\}, "required": ["context"]\}\}\}

</tools>

For each function call, return a json object with function name and arguments within <tool\_call></tool\_call> XML tags:

<tool\_call>

\{"name": <function-name>, "arguments": <args-json-object>\}

</tool\_call>
}
\caption{\textbf{System prompt framing summary as a tool.}}
\label{fig:promptsystem}
\end{figure}

\begin{figure}
\centering
\displaycode{
<|im\_start|>user

\{problem\}

~

Context:

\{context\}<|im\_end|>

<|im\_start|>assistant
}
\caption{\textbf{User prompt inserting context simply.}}
\label{fig:promptusercontext}
\end{figure}

\begin{figure}
\centering
\displaycode{
\{\autoref{fig:promptsystem}\}

~

Below is an example of using the pass tool:

<|im\_start|>user

There exist real numbers $x$ and $y$, both greater than 1, such that $\log_x\left(y^x\right)=\log_y\left(x^{4y}\right)=10$. Find $xy$.<|im\_end|>

<|im\_start|>assistant

<think>

Okay, so I need to solve this problem where there are real numbers x and y, both greater than 1, such that log base x of ($y^x$) equals log base y of ($x^4y$) equals 10. And I need to find the product xy. Hmm, let me think.

~

First, let me recall some logarithm properties. The equation says that two different logarithms are equal to 10. So, $\log_x(y^x)$ = 10 and $\log_y(x^{4y})$ = 10. Maybe I can convert these logarithmic equations into exponential form to make them easier to handle.

~

Starting with the first one: $\log_x(y^x) = 10$. Using the logarithm power rule, which says that $\log_b(a^c) = c*\log_b(a)$. So, applying that here, $\log_x(y^x) = x * \log_x(y) = 10$. So, x * $\log_x(y)$ = 10.

~

But $\log_x(y)$ is the same as $1 / \log_y(x)$, by the change of base formula. Wait, but maybe I can express $\log_x(y)$ in terms of something else. Let me

<tool\_call>

\{"name": "pass", "arguments": \{"context": "Maybe convert logarithmic equations into exponential form to make them easier to handle. Via logarithm power rule $\log_x(y^x) = x * \log_x(y) = 10$. Noting that $\log_x(y)$ is the same as $1 / \log_y(x)$. Maybe try express $\log_x(y)$ in terms of something else."\}\}

</tool\_call>

~

When context is passed to you, evaluate its correctness first and then use it if correct.

<|im\_start|>user

There exist real numbers $x$ and $y$, both greater than 1, such that $\log_x\left(y^x\right)=\log_y\left(x^{4y}\right)=10$. Find $xy$.<|im\_end|>

<|im\_start|>assistant

<context>

Maybe convert logarithmic equations into exponential form to make them easier to handle. Via logarithm power rule $\log_x(y^x) = x * \log_x(y) = 10$. Noting that $\log_x(y)$ is the same as $1 / \log_y(x)$. Maybe try express $\log_x(y)$ in terms of something else.

</context>

<think>

Okay, so I need to first evaluate whether the context is correct and then use it to solve the problem.

~

The context suggests converting logarithmic equations into exponential form and using the logarithm power rule, which is a valid approach. It also notes that $\log_x(y)$ can be expressed in terms of $\log_y(x)$, which is correct. It suggests expressing $\log_x(y)$ in terms of something else.

~

Wait, do we really need to express $\log_x(y)$ in terms of something else? Maybe we can instead first use the logarithm power rule for the second equation as well. Let me do that...
}
\caption{\textbf{System prompt with tool-usage and context-usage examples.}}
\label{fig:promptsystemshot}
\end{figure}

\begin{figure}
\centering
\displaycode{
<|im\_start|>user

\{problem\}

~

Context from your prior reasoning attempt that may be useful but could also be wrong:

\{context\}<|im\_end|>

<|im\_start|>assistant
}
\caption{\textbf{User prompt inserting context with extra info.}}
\label{fig:promptuserwarning}
\end{figure}

\FloatBarrier

\subsection{\texorpdfstring{$\mathbf{H_2O}$}{H2O}}

$H_2O$~\citep{zhang2023h2oheavyhitteroracleefficient} retains recent tokens and ``heavy hitters'' that accumulate high attention scores, while evicting less influential tokens from the KV cache. Thus, prefix tokens that have not received sufficiently high attention can be discarded once they leave the recency window. $H_2O$ is not integrated with FlashAttention/vLLM (\href{https://github.com/vllm-project/vllm/issues/3532}{https://github.com/vllm-project/vllm/issues/3532}), making a fair efficiency comparison difficult.

However, we think combining $H_2O$ with \method could be promising. Both retain recent tokens, but they differ in that \method is ``forward-looking'' while $H_2O$ is ``backward-looking''. \method preserves prefix tokens, which we know may become important later. For example, a tool definition may get little attention for thousands of tokens but must be available when the model eventually uses that tool. In contrast, $H_2O$ retains tokens based on their importance thus far. Preserving the prefix, heavy-hitting intermediate tokens, and a recency window may outperform \method alone. However, it may be challenging to make such a method work for reinforcement learning.

\section{Analyzing outputs}

\begin{figure}[h!]
\centering
\displaycode{
<|im\_start|>user

An isosceles trapezoid has an inscribed circle tangent to each of its four sides. The radius of the circle is $3$, and the area of the trapezoid is $72$. Let the parallel sides of the trapezoid have lengths $r$ and $s$, with $r \neq s$. Find $r^2+s^2$

Context from your prior reasoning attempt that may be useful but could also be wrong:

I need to compute r² + s² where $r = 12 + 6\sqrt{3}$ and $s = 12 - 6\sqrt{3}$.<|im\_end|>

<|im\_start|>assistant

<think>

Okay, so I need to find r² + s² where r and s are the lengths of the two bases of an isosceles trapezoid that has an inscribed circle with radius 3 and an area of 72. Hmm, let me recall what I know about trapezoids with inscribed circles.

First, I remember that for a trapezoid to have an inscribed circle (be tangential), the sum of the lengths of the two bases must equal the sum of the lengths of the two legs. Since it's an isosceles trapezoid, the legs are equal in length. Let me note that down:

Let the two bases be r and s, with $r \neq s$. Then, the legs (which are equal) must satisfy:

r + s = 2 * leg\_length.

But I also know that the area of a trapezoid is given by the formula:
}
\caption{\textbf{Summary duplicate processing problem on AIME sample.} The model ignores its prior derivations and starts the task anew.}
\label{fig:summissue}
\end{figure}

\begin{figure}[h!]
\centering
\displaycode{
<|im\_start|>user

An isosceles trapezoid has an inscribed circle tangent to each of its four sides. The radius of the circle is $3$, and the area of the trapezoid is $72$. Let the parallel sides of the trapezoid have lengths $r$ and $s$, with $r \neq s$. Find $r^2+s^2$<|im\_end|>

<|im\_start|>assistant

<think>

$+ s)/2 * h = 24/2 * 6 = 12 * 6 = 72$. Correct. The radius is 3, which is $h/2 = 6/2 = 3$. Correct.

\colorbox{s1bluelight}{Then, r² + s²: $(12 - 6\sqrt{3})^2 + (12 + 6\sqrt{3})^2$.}

\colorbox{s1bluelight}{Let me compute $(a - b)^2 + (a + b)^2$ where $a = 12, b = 6\sqrt{3}$.}

Which is (a² - 2ab + b²) + (a² + 2ab + b²) = 2a² + \colorbox{yellow}{2}b² = 2(a² + b²).

So $2*(12^2 + (6\sqrt{3})^2) = 2*(144 + 36*3) = 2*(144 + 108) = 2*(252) = 504$.

Wait, but the answer is 504? But let me check again.

Wait, the problem says the trapezoid has an inscribed circle. For a trapezoid to have an inscribed circle (be tangential), the sum of the lengths of the two bases must equal the sum of the lengths of the two legs. So, in this case, since it's an isosceles trapezoid, the legs are equal. So, let me check if that's the case.

Let me recall that for a trapezoid to be tangential, the sum of the lengths of the two bases equals the sum of the lengths of the two legs. So, if the bases are r and s, and the legs are both of length l, then r + s = 2l.

Also, the area is given by the average of the bases times the height. So, (r + s)/2 * h = 72. And the radius of the inscribed circle is h/2 = 3, so h = 6. Therefore, (r + s)/2 * 6 = 72 => (r + s)/2 = 12 => r + s = 24.

So, we have that r + s = 24 and that r + s = 2l => l = 12. So, the legs are 12 each.

Now, the trapezoid is isosceles, so the legs are equal, and the height is 6.

Now, we need to find r² + s². But how can we find r and s? We know that r + s = 24, and we need another equation. Let's think about the geometry.

In an isosceles trapezoid with bases r and s, and legs of length l = 12, and height h = 6.

The difference between the bases is r - s, and when you drop the perpendiculars from the ends of the shorter base to the longer base, you form two right triangles on either side. The base of each triangle is (r - s)/2, and the height is 6, and the hypotenuse is l = 12.

So, (r - s)/2 = sqrt(l² - h²) = sqrt(144 - 36) = sqrt(108) = 6*sqrt(3).

Therefore, $(r - s)/2 = 6\sqrt{3} => r - s = 12\sqrt{3}$.

So now we have:

r + s = 24

$r - s = 12\sqrt{3}$

Adding these equations: $2r = 24 + 12\sqrt{3} => r = 12 + 6\sqrt{3}$

Subtracting the second equation from the first:

$2s = 24 - 12\sqrt{3} => s = 12 - 6\sqrt{3}$

Therefore, r² + s² = $(12 + 6\sqrt{3})^2 + (12 - 6\sqrt{3})^2$

\colorbox{s1bluelight}{Let me compute each term:}

First term: $(12 + 6\sqrt{3})^2 = 12^2 + 2*12*6\sqrt{3} + (6\sqrt{3})^2 = 144 + 144\sqrt{3} + 36*3 = 144 + 144\sqrt{3} + 108 = 252 + 144\sqrt{3}$

Second term: $(12 - 6\sqrt{3})^2 = 12^2 - 2*12*6\sqrt{3} + (6\sqrt{3})^2 = 144 - 144\sqrt{3} + 108 = 252 - 144\sqrt{3}$

Adding them together: $(252 + 144\sqrt{3}) + (252 - 144\sqrt{3}) = 252 + 252 = 504$

So the answer is 504. Therefore, the value of r² + s² is 504.
}
\caption{\textbf{Last k duplicate processing issue on AIME sample.} We mark the \colorbox{s1bluelight}{repetition} and where the carried-over 256 tokens \colorbox{yellow}{end}.}
\label{fig:lastkissue}
\end{figure}

\begin{figure}[h!]
\centering
\displaycode{
...

Let me think of the following code:

def can\_remove\_k(targetIndices, source, pattern, k):

~~~~removed = []

~~~~for i in range(len(targetIndices)):

~~~~~~~~if len(removed) >= k:

~~~~~~~~~~~~break

~~~~~~~~\# Check if removing this index allows the pattern to be a subsequence

~~~~~~~~\# Simulate the pattern as a subsequence, skipping this index

~~~~~~~~\# So, we need to check if the pattern is a subsequence after removing this index

~~~~~~~~\# To do this, we can simulate the pattern as a subsequence, but skipping this index

~~~~~~~~\# Let's create a list of indices in source that are not removed

~~~~~~~~\# But since the targetIndices are in order, and we are checking for each index in order, we can simulate this

~~~~~~~~\# So, we can simulate the pattern as a subsequence, with the current index being removed

~~~~~~~~\# So, we can create a list of the source's characters, and for each character, if it's not the current index, we check

~~~~~~~~\# But this is O(n) per check

~~~~~~~~\# So, the code would be:

~~~~~~~~\# simulate the pattern as a subsequence, with the current index removed

~~~~~~~~\# if the pattern is a subsequence, then we can remove it

~~~~~~~~\# So, we need to check if the pattern is a subsequence of source, with the current index removed

~~~~~~~~\# Let's implement this

~~~~~~~~\# We can create a list of the source's characters, and for each character, if it's not the current index, we check

~~~~~~~~\# But how to do this efficiently

~~~~~~~~\# Let's create a copy of the source, and remove the current index

~~~~~~~~\# But this is O(n) for each check

~~~~~~~~\# However, for large n, this is not feasible

~~~~~~~~\# So, we need a more efficient way to check this

~~~~~~~~\# Here's a way to do it:

~~~~~~~~\# We can simulate the pattern as a subsequence, but with the current index removed

~~~~~~~~\# So, we can have a pointer for the pattern, and iterate through the source, skipping the current index

~~~~~~~~\# So, the code would be:

~~~~~~~~\# current pattern pointer

~~~~~~~~p = 0

~~~~~~~~for idx in range(len(source)):

~~~~~~~~~~~~if idx == targetIndices[i]:

~~~~~~~~~~~~~~~~continue

~~~~~~~~~~~~if p < len(pattern) and source[idx] == pattern[p]:

~~~~~~~~~~~~~~~~p += 1

~~~~~~~~if p == len(pattern):

~~~~~~~~~~~~removed.append(targetIndices[i])

~~~~return len(removed) >= k

Wait, but this code is not correct. Because for each check, we are trying to remove the current index, but in the code above, we are checking whether the pattern is a subsequence after removing the current index.

...
}
\caption{\textbf{Reasoning in comments issue on LiveCodeBench sample.}}
\label{fig:lcb}
\end{figure}

\end{document}